\documentclass{naturep}

\usepackage{amssymb}
\usepackage{amsmath}
\usepackage{graphicx}
\usepackage{algorithmicx}
\usepackage{algorithm}
\usepackage{subfigure}
\usepackage{subcaption}
\usepackage[noend]{algpseudocode}
\usepackage{bbm}
\usepackage[margin=0.6in]{geometry}
\usepackage{ragged2e}
\usepackage{setspace}
\usepackage{longtable}
\usepackage{hyperref}
\usepackage{anyfontsize}
\usepackage{setspace}
\usepackage{float}
\usepackage{booktabs}
\usepackage{multirow}
\usepackage{xcolor}
\usepackage{blindtext}
\usepackage{tabularx}
\usepackage{caption}
\usepackage{array}
\usepackage{bbding}
\usepackage{pifont}
\usepackage{fontawesome}

\usepackage[normalem]{ulem}

\newcommand{\cmark}{\ding{51}}
\newcommand{\xmark}{\ding{55}}

\usepackage{subcaption}

\makeatletter
\let\saved@includegraphics\includegraphics
\AtBeginDocument{\let\includegraphics\saved@includegraphics}

\makeatother

\usepackage{enumitem, kantlipsum}

\title{MAP: Evaluation and Multi-Agent Enhancement of Large \\Language Models for Inpatient Pathways}

\begin{document}
% \linenumbers

\maketitle
\begin{spacing}{1.8}
% \vspace{-15mm}
\vspace{-10mm}
\noindent Zhen Chen$^{1\boldsymbol{\ddag}}$, Zhihao Peng$^{1\boldsymbol{\ddag}}$, Xusheng Liang$^{1\boldsymbol{\ddag}}$, Cheng Wang$^{1}$, Peigan Liang$^{2}$, Linsheng Zeng$^{3}$, Minjie Ju$^{4}$, Yixuan Yuan$^{1*}$
\end{spacing}
\vspace{-10mm}
\begin{spacing}{1.4}
\begin{affiliations}
 \item Department of Electronic Engineering, The Chinese University of Hong Kong, Hong Kong
 % \item Centre for Artificial Intelligence and Robotics, Chinese Academy of Sciences, Hong Kong
 \item Guangzhou Hospital of Integrated Traditional Chinese and Western Medicine, Guangzhou, China
 \item Shenzhen Traditional Chinese Medicine Hospital, Shenzhen, China
 \item Department of Critical Care Medicine, Zhongshan Hospital Fudan University, China
 \\$\boldsymbol{\ddag}$ Contributed Equally
 \\\textbf{*Corresponding author}: Yixuan Yuan (yxyuan@ee.cuhk.edu.hk)
\end{affiliations}
\end{spacing}

\vspace{-5mm}
\begin{spacing}{1.0}
\section{Abstract} 

Inpatient pathways demand complex clinical decision-making based on comprehensive patient information, posing critical challenges for clinicians. Despite advancements in large language models (LLMs) in medical applications, limited research focused on artificial intelligence (AI) inpatient pathways systems, due to the lack of large-scale inpatient datasets. Moreover, existing medical benchmarks typically concentrated on medical question-answering and examinations, ignoring the multifaceted nature of clinical decision-making in inpatient settings. To address these gaps, we first developed the Inpatient Pathway Decision Support (IPDS) benchmark from the MIMIC-IV database, encompassing 51,274 cases across nine triage departments and 17 major disease categories alongside 16 standardized treatment options. Then, we proposed the Multi-Agent Inpatient Pathways (MAP) framework to accomplish inpatient pathways with three clinical agents, including a triage agent managing the patient admission, a diagnosis agent serving as the primary decision maker at the department, and a treatment agent providing treatment plans. Additionally, our MAP framework includes a chief agent overseeing the inpatient pathways to guide and promote these three clinician agents. Extensive experiments showed our MAP improved the diagnosis accuracy by 25.10\% compared to the state-of-the-art LLM HuatuoGPT2-13B. It is worth noting that our MAP demonstrated significant clinical compliance, outperforming three board-certified clinicians by 10\%-12\%, establishing a foundation for inpatient pathways systems.
\end{spacing}

\newpage

\begin{figure*}[!]
\centering
   \includegraphics [width=0.98 \textwidth]{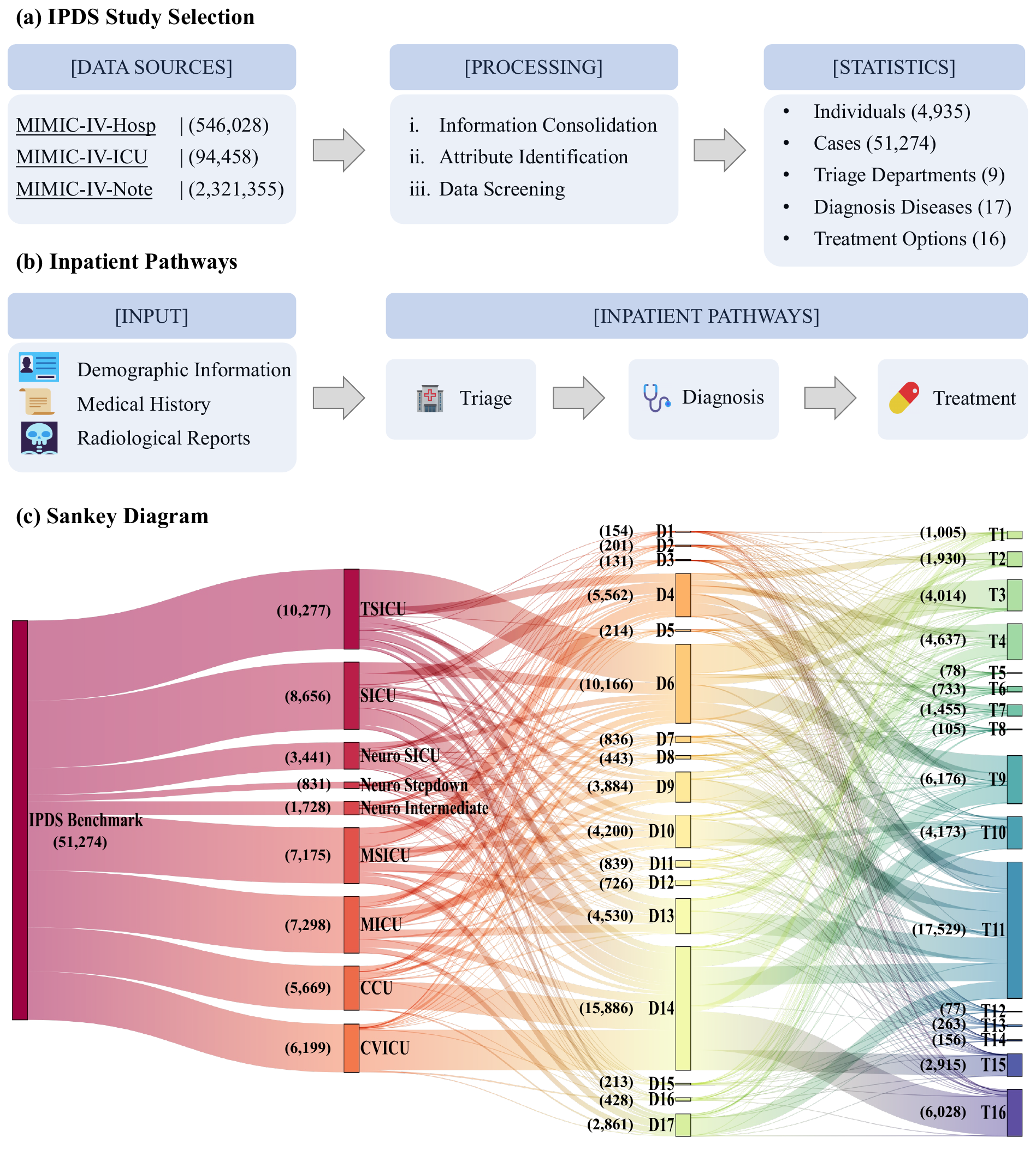}
\caption{\textbf{Illustration of the Inpatient Pathway Decision Support (IPDS) benchmark.} (a) The statistics, processing, and data sources of the IPDS benchmark. The IPDS contains 51,274 cases across 9 departments, 17 diseases (D1-D17), and 16 treatments (T1-T16), and provides the comprehensive evaluation of LLMs in different inpatient scenarios. The specific abbreviations and details of the diagnosis options were detailed in Table \ref{tab: dise_table}. (b) The evaluation of the IPDS benchmark. (c) The Sankey diagram of the IPDS benchmark. This diagram visualizes the data distribution in the workflow of different inpatient scenarios. The specific abbreviations and details of the department, disease, and treatment options are provided in the supplementary materials.}
\label{fig: benchmark}
\end{figure*}

\begin{figure*}[h!]
\centering
   \includegraphics [width=0.92 \textwidth]{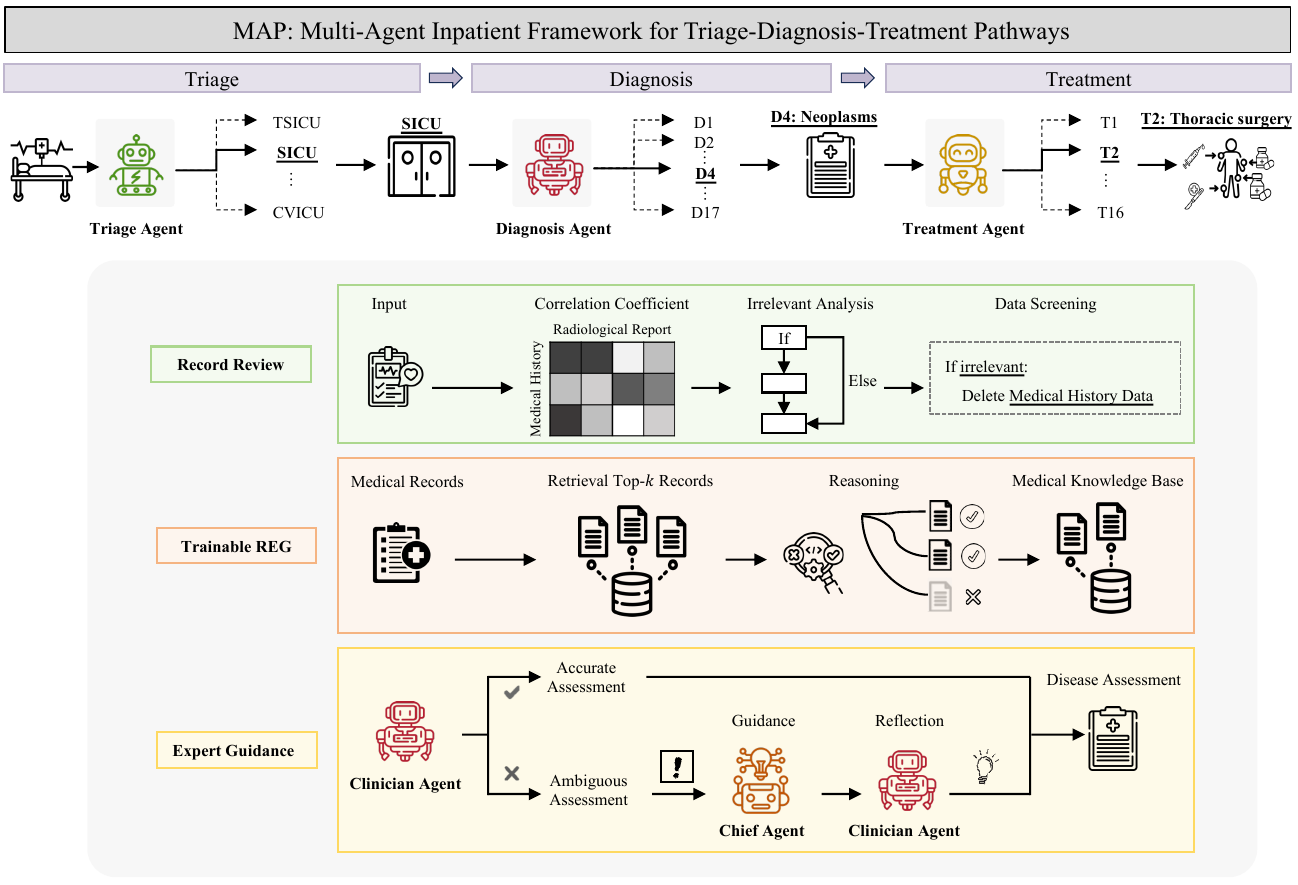}
\caption{\textbf{Overview of the Multi-Agent Inpatient Pathways (MAP) framework.} The MAP is a multi-agent collaborative framework that simulates the inpatient pathway flow. The framework consists of LLM-empowered agents: a triage agent for department triage, a diagnosis agent for each department for the clinical decision-maker, a treatment agent for the treatment plan, and a chief agent for overseeing diagnosis and treatment pathways. Three key components support our MAP framework: (1) a record review module that analyzes patient data, including demographic information, radiological reports, and medical history; (2) a trainable REG module that integrates clinical knowledge bases with chain-of-thought reasoning to support reliable diagnostic decision-making; and (3) an expert guidance module that ensures diagnostic rigor through structured supervision of the diagnosis agent.}
\label{fig: MAP}
\end{figure*}

\section{Introduction}

As a critical component of worldwide healthcare systems, inpatient pathways \cite{alagappan2007international,topol2019high,rajpurkar2022ai,avram2024ai,hager2024evaluation,van2024adapted,schoenfelder2011determinants,dixon2012exome,xu2024natu,varghese2024artificial} are characterized by the necessity for clinical decision-making based on comprehensive patient information, and present challenges for clinicians across all levels of experience. Even experienced clinicians encounter challenges in the inpatient pathways, where the complexity of clinical decision-making in time-sensitive situations can affect diagnostic accuracy. It is suggested that diagnostic errors lead to 40,000 to 80,000 fatalities annually, affecting more than 250,000 Americans who encounter such errors while receiving care in hospitals in the United States \cite{graber2017impact}. The consequences of diagnostic errors can lead to a cascade of adverse outcomes, ranging from unnecessary treatments and prolonged hospital stays to disability or even death, highlighting the critical importance of reliable clinical decision support systems in healthcare \cite{lauritsen2020explainable, erion2022coai, underdahl2024physician, ong2024artificial, chen2024multimodal}. With the breakthrough of large language models (LLMs), recent studies have demonstrated promising capabilities in medical knowledge retrieval \cite{wang2024prompt,preiksaitis2023chatgpt}, consultation systems \cite{wu2024clinical,moor2023foundation}, and diagnostic suggestions \cite{zhou2024pre,peng2024gbt,kanithi2024medic,yang2024limits}. LLMs can leverage publicly available medical knowledge to enhance medical reasoning and comprehension, demonstrating the potential to rival clinicians in medical licensing examinations \cite{gilson2023does}. Additionally, LLMs can translate complex medical terminology into plain words, making physician-patient interactions more efficient by clearly explaining medical procedures and treatment options \cite{johri2025evaluation}. However, the effectiveness of LLMs in supporting inpatient pathways still remains unexplored. Especially, the complex inpatient pathways demand sophisticated capabilities, including the systematic analysis of diverse and complex clinical information from electronic health records, prioritization of critical issues within the dynamic and evolving nature of inpatient settings, and support for nuanced diagnostic decision-making within these complex clinical environments. 

\noindent To fill these gaps, we developed the Inpatient Pathway Decision Support (IPDS) benchmark by systematically integrating comprehensive clinical data of the Medical Information Mart for Intensive Care (MIMIC) database at the Beth Israel Deaconess Medical Center \cite{johnson2016mimic,pollard2018eicu,hyland2020early,zeng2020pic,thoral2021sharing,johnson2023mimic,johnsonmimic}. Unlike existing medical benchmarks \cite{jin2021disease,pal2022medmcqa,wornow2023ehrshot,cai2024medbench,wu2024clinical,bae2024ehrxqa,chen2024multimodal} that primarily focused on medical licensing exams and general clinical questions, the IPDS benchmark encompasses 51,274 patient cases across 9 clinical departments, 17 disease categories, and 16 standardized treatment options. Our data pipeline utilized an international disease statistical classification list \cite{world2004international} to recategorize 1,298 original disease labels into 17 broader categories. As such, the IPDS incorporates curated demographic information, radiological reports, and medical history, which are essential for comprehensive inpatient pathway decision support, providing a comprehensive representation of inpatient scenarios, as shown in Figure \ref{fig: benchmark}. We empirically observed that state-of-the-art LLMs merely achieved unsatisfying performance on inpatient scenarios of the IPDS, e.g., the general LLaMA-3-8B with 49.30\%, InternLM2-20B with 51.70\%, and the medical HuatuoGPT2-13B with 53.00\% in the accuracy of the diagnosis task. These findings indicate that the state-of-the-art LLMs can hardly meet the requirements of inpatient scenarios, resulting in an urgent need for a diagnosis support framework with significantly improved inpatient pathway performance.

\noindent We then developed the Multi-Agent Inpatient Pathways (MAP) framework to support the Triage-Diagnosis-Treatment (TDT) clinical pathway, as illustrated in Figure \ref{fig: MAP}. Specifically, our MAP framework established the collaboration among three clinician agents and a chief agent, including the triage agent that manages the patient admission, the diagnosis agent that acts as the primary decision-maker in the department, the treatment agent that provides treatment plans, and the chief agent that oversees different clinical tasks along inpatient pathways. To promote collaboration among these agents, our MAP was elaborately developed with three modules. The record review module employed a semantic analysis component to comprehend medical terminology and clinical descriptions of patient data. The trainable retrieval-enhanced generation (REG) module retrieves the most relevant medical records from an extensive knowledge base, simulating the case review process of the diagnosis agent to maintain diagnostic accuracy. The expert guidance module achieves the supervisory relationship between the diagnosis agent and the supervisor chief agent of each department. The effectiveness of these modules was verified by the ablation studies in the supplementary materials.

\noindent We conducted extensive studies on supporting inpatient pathways by comparing our MAP and state-of-the-art LLMs \cite{chen2023huatuogpt,chen2023meditron,toma2023clinical,cai2024internlm2,dubey2024llama} across three distinct inpatient scenarios of TDT. First, the \textbf{triage} task determines the most suitable department for patients based on symptoms, medical history, and urgency, involving high-level categorization into broad clinical areas.
After that, the \textbf{diagnosis} task focuses on utilizing diagnostic information (e.g., demographic information, radiological reports, and medical history) to identify specific diseases or conditions, requiring in-depth analysis of clinical data. 
Finally, the \textbf{treatment} task aims to select proper treatment options considering the patient's specific needs and condition severity, focusing on optimizing outcomes and minimizing risks, involving decision-making about interventions and follow-up care.
Our findings indicate that the proposed MAP framework significantly enhances the capabilities of LLMs in supporting inpatient pathways, particularly in complex disease cases where state-of-the-art LLMs exhibited unsatisfactory performance. Furthermore, we examined our MAP and state-of-the-art LLMs concerning diagnostic challenges across various disease categories, analyzed the conditions leading to misdiagnosis, and evaluated diagnostic performance while excluding specific inputs. We further randomly sampled 100 cases from the IPDS test set and conducted the statistical analysis of clinicians' diagnostic performance, assessing the clinical compliance of our MAP and advanced LLMs with expert judgments. In summary, our MAP framework, benefiting from a diagnostic architecture specifically designed for the inpatient pipeline, improves the AI system to provide more accurate and effective inpatient support. We have made our IPDS and MAP publicly available to facilitate further research in inpatient diagnostic support.

\begin{table*}[]
\caption{\textbf{Statistics comparison of existing benchmarks and our IPDS benchmark, including the source type, number of samples, and multiple departments involvement.} In general, the main shortcomings of existing evaluation benchmarks include ($i$) lack of comprehensive and evenly distributed departmental coverage to prevent evaluation bias; ($ii$) data sources often come from easily accessible online consultation platforms, medical textbooks, and professional examinations, which poses high risks of data leakage; ($iii$) existing benchmarks primarily test medical knowledge through multiple-choice questions, which differ significantly from real-world diagnostic scenarios.}
\label{tab: benchmarks}
\centering
\resizebox{0.8\linewidth}{!}{
\begin{tabular}{cccccc}
\hline\hline
Benchmark                                                 & Data Source & Samples & Multiple Departments      \\ \hline
PubMedQA \cite{jin2019pubmedqa} [EMNLP'19]                & Public      & 500     & \textcolor{red}{\xmark}   \\
MedQA \cite{jin2021disease} [ASci.'21]                    & Public      & 1,273   & \textcolor{red}{\xmark}   \\
CMB-Clin \cite{wang2024cmb} [NAACL'24]                    & Public      & 74      & \textcolor{red}{\xmark}   \\
HealthSearchQA \cite{singhal2023large} [Nature'23]        & Public      & 3,173   & \textcolor{red}{\xmark}   \\
CMExam \cite{liu2024benchmarking} [NeurIPS'24]            & Public      & 68,119  & \textcolor{green}{\cmark} \\
MedBench \cite{cai2024medbench} [AAAI'24]                 & Private     & 1,025   & \textcolor{red}{\xmark}   \\
ClinicalBench \cite{yan2024clinicallab} [ArXiv'24]        & Private     & 1,500   & \textcolor{green}{\cmark} \\
MIMIC-CDM \cite{hager2024evaluation} [Nature Medicine'24] & Public      & 2,400   & \textcolor{red}{\xmark}   \\ \hline
IPDS (Ours)                                               & Public      & 51,274  & \textcolor{green}{\cmark} \\ \hline\hline
\end{tabular}
}
\end{table*}

\section{Benchmark} 
To address the critical need for evaluating LLMs in supporting inpatient pathways, we introduced the IPDS, a novel benchmark systematically integrated from the MIMIC-IV database, specifically from MIMIC-IV-Hosp, MIMIC-IV-ICU, and MIMIC-IV-Note databases of the Beth Israel Deaconess Medical Center \cite{johnson2016mimic,johnson2023mimic,johnsonmimic}.
The IPDS encompasses 51,274 patient cases across 9 departments, 17 major disease categories, and 16 standardized treatment pathway options, as shown in Figure \ref{fig: benchmark}. We compared our IPDS with existing clinical benchmarks in Table \ref{tab: benchmarks}, in terms of the source type, number of samples, and multiple-department involvement. Unlike state-of-the-art medical benchmarks \cite{alerhand2017time,carr2010defining,sklar2010future,france2006system}, the IPDS is elaborately designed for inpatient diagnostic support, incorporates comprehensive patient data (\textit{e.g.}, demographic information, radiological reports, and medical history) to complete three distinct clinical classification tasks, including the triage, diagnosis, and treatment pathways. Specifically, the triage task is simplified as a nine-category classification task aiming to correctly evaluate the triage agent's ability to direct patients to the appropriate department. The diagnosis task is a multi-class classification task with seventeen disease categories, which is a critical bridge between the initial department assignment and the subsequent treatment pathway in inpatient care. The treatment task is a multi-category classification task with sixteen treatment options designed to identify whether the patient is receiving the appropriate treatment. To this end, the multifaceted structure of the IPDS benchmark allows for a thorough assessment of LLM capabilities in three critical areas, including medical information analysis, consolidation, and diagnostic decision-making, which is essential to inpatient diagnostic support. The development of the IPDS benchmark followed rigorous de-identification protocols \cite{neamatullah2008automated,johnson2020deidentification} and access guidelines \cite{morley2020ethics,elendu2023ethical} to ensure patient privacy protection and data integrity. These measures made the IPDS a valuable and responsible resource for advancing LLMs in inpatient diagnostic support. We split 1,000 samples of IPDS to evaluate the model’s ability to support inpatient pathways for triage, diagnosis, and treatment tasks, and the remaining samples were used for model training. Beyond this evaluation, the IPDS aims to facilitate the development of LLMs specifically designed for supporting inpatient pathways, establishing new standards for creating and assessing clinical decision support systems. 

\begin{figure*}[!]
\centering
    \subfigure[Triage]{
    \includegraphics [width=0.280 \textwidth]{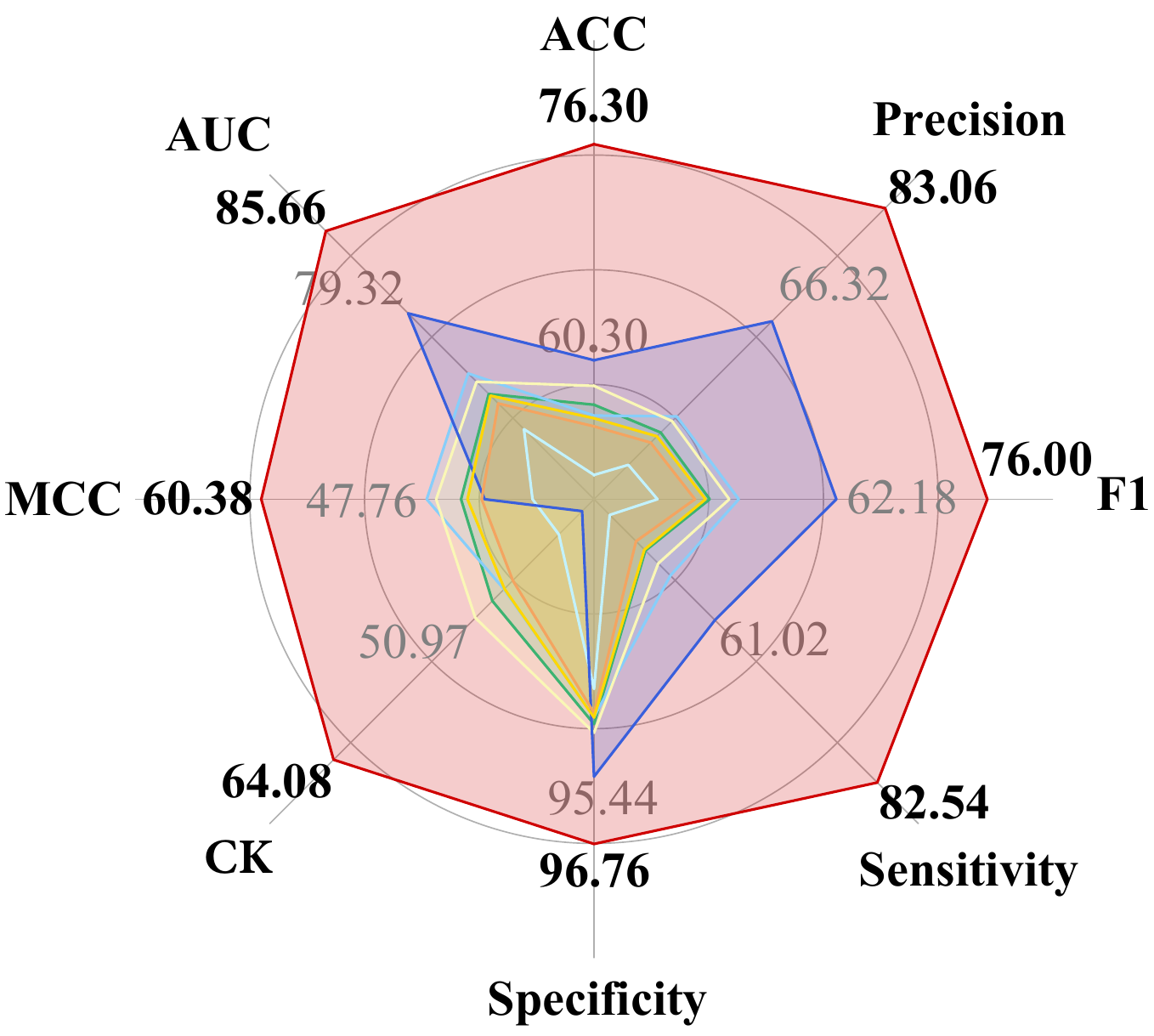} % width=0.480 \textwidth
    }
    \subfigure[Diagnosis]{
    \includegraphics [width=0.280 \textwidth]{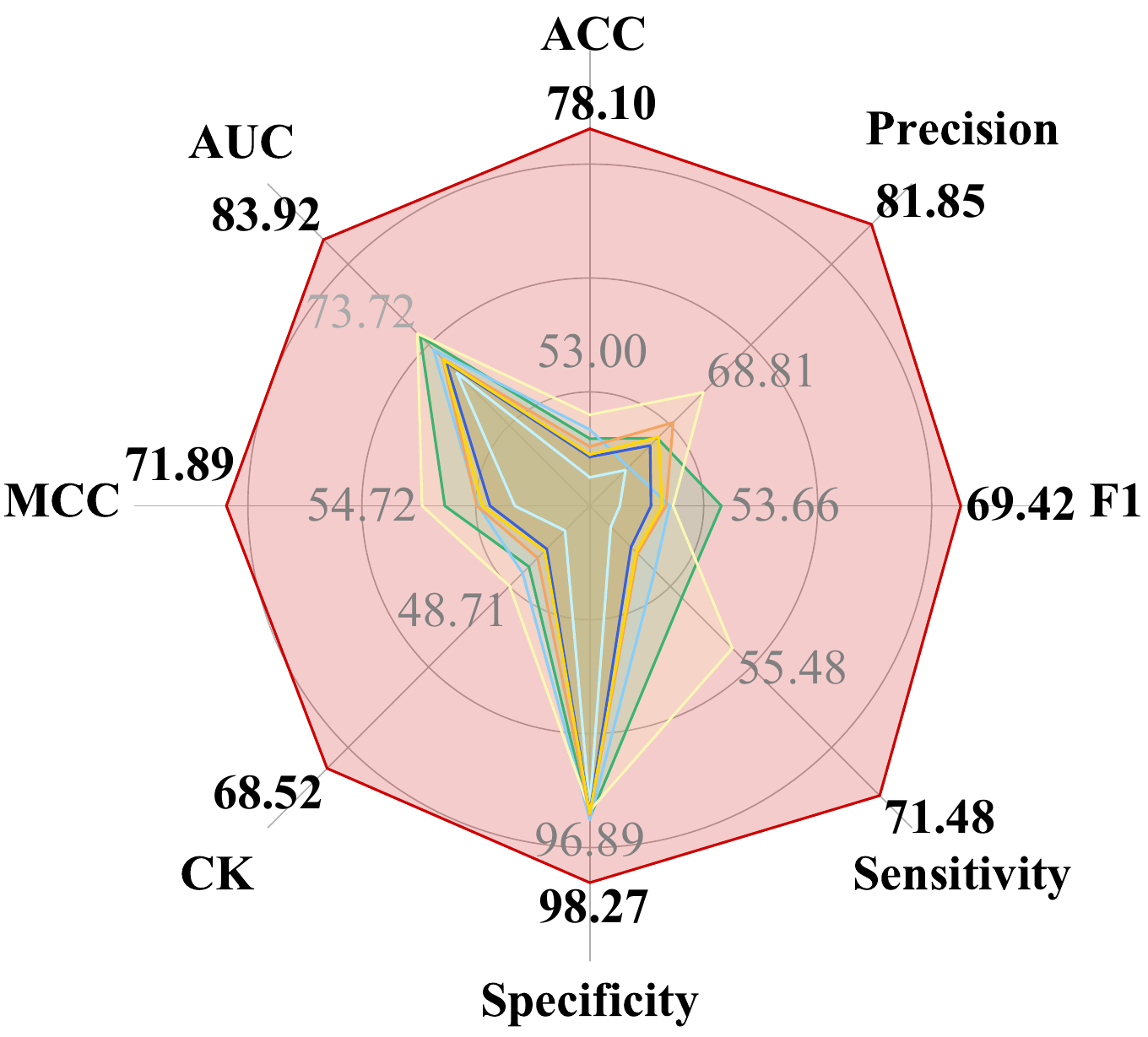}
    }
    \subfigure[Treatment]{
    \includegraphics [width=0.36 \textwidth]{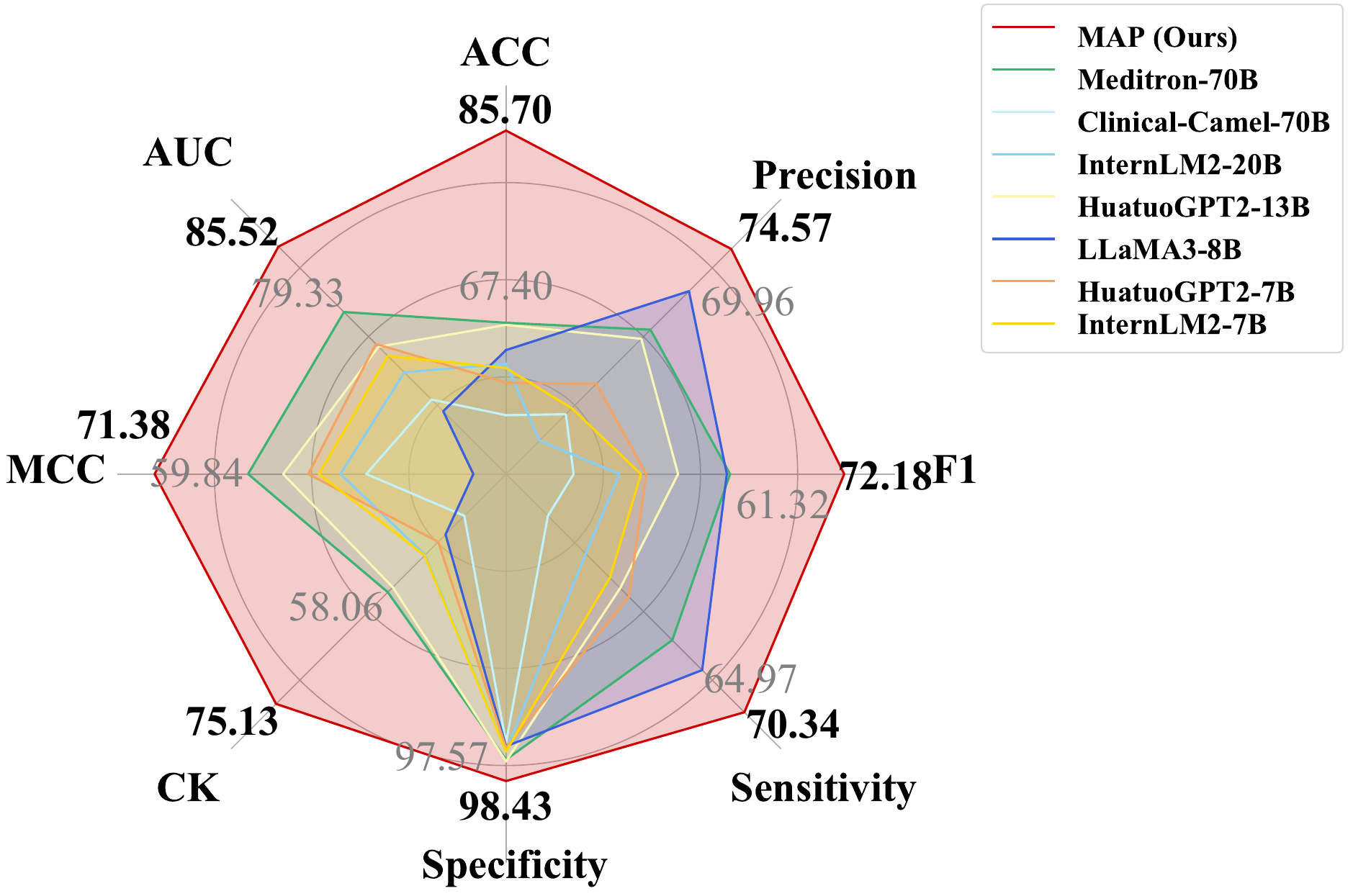}
}
\caption{\textbf{The MAP demonstrated the enhanced capabilities in supporting inpatient pathways compared to state-of-the-art LLMs.} Current general LLMs showed limited performance in diagnostic support, with an accuracy below 53.00\%, while specialized medical models such as Clinical-Camel-70B achieve 47.50\% accuracy. The MAP achieved an overall diagnostic support accuracy of 78.10\%, showing improvements of 30.60\%, 27.20\%, and 25.10\% over Clinical-Camel-70B, Meditron-70B, and HuatuoGPT2-13B ($p<0.001$ for all comparisons). The best and second-best performance values are marked for better clarity. These results demonstrated the potential of the MAP as a clinical decision-support tool across diverse inpatient scenarios.}
\label{fig: Dise_overall_each}
\end{figure*}

\section{Results}
To assess the capability of varying LLMs in supporting inpatient pathways, we evaluated the general LLMs (InternLM2-7B/20B \cite{cai2024internlm2}, LLaMA-3-8B \cite{dubey2024llama}), the specialized medical LLMs (HuatuoGPT2-7B/13B \cite{zhang2023huatuogpt}, Clinical-Camel-70B \cite{toma2023clinical}, Meditron-70B \cite{chen2023meditron}), and our MAP framework. Due to privacy concerns and data agreements\footnote{\href{https://physionet.org/about/licenses/physionet-credentialed-health-data-license-150/}{https://physionet.org/about/licenses/physionet-credentialed-health-data-license-150/}}, the MIMIC database prohibits the use of its data with external APIs such as OpenAI or Google, preventing the IPDS research from evaluating ChatGPT \cite{floridi2020gpt}, GPT-4 \cite{achiam2023gpt}, and Med-PaLM \cite{singhal2023towards}. Our systematic studies reveal several significant findings regarding the capabilities and limitations of LLMs in supporting inpatient pathways.

\subsection{The under-performance of state-of-the-art LLMs in supporting inpatient pathways.}

\noindent
Our analysis revealed significant limitations in the ability of state-of-the-art LLMs to support inpatient pathways across triage, diagnosis, and treatment tasks. As illustrated in Figure \ref{fig: Dise_overall_each}, we compared the classification performance of our MAP and state-of-the-art LLMs on these three common clinical classification tasks, where we can find that state-of-the-art LLMs have demonstrated unsatisfying performance in inpatient scenarios within the IPDS. For example, LLaMA-3-8B achieved only 60.30\%, 49.30\%, and 64.80\% accuracy in triage, diagnosis, and treatment inpatient pathway tasks, respectively. Similar limitations exist in specialized medical LLMs with HuatuoGPT2-13B (58.40\%, 53.00\%, 67.20\%), Clinical-Camel-70B (51.80\%, 47.50\%, 58.60\%), and Meditron (57.00\%, 50.90\%, 67.40\%). These results revealed the substantial potential for improvement in supporting the inpatient diagnostic capabilities of LLMs.

\begin{figure*}[t]
\centering
    \subfigure[Specific Disease Analyses]{
    \includegraphics [width=0.38 \textwidth]{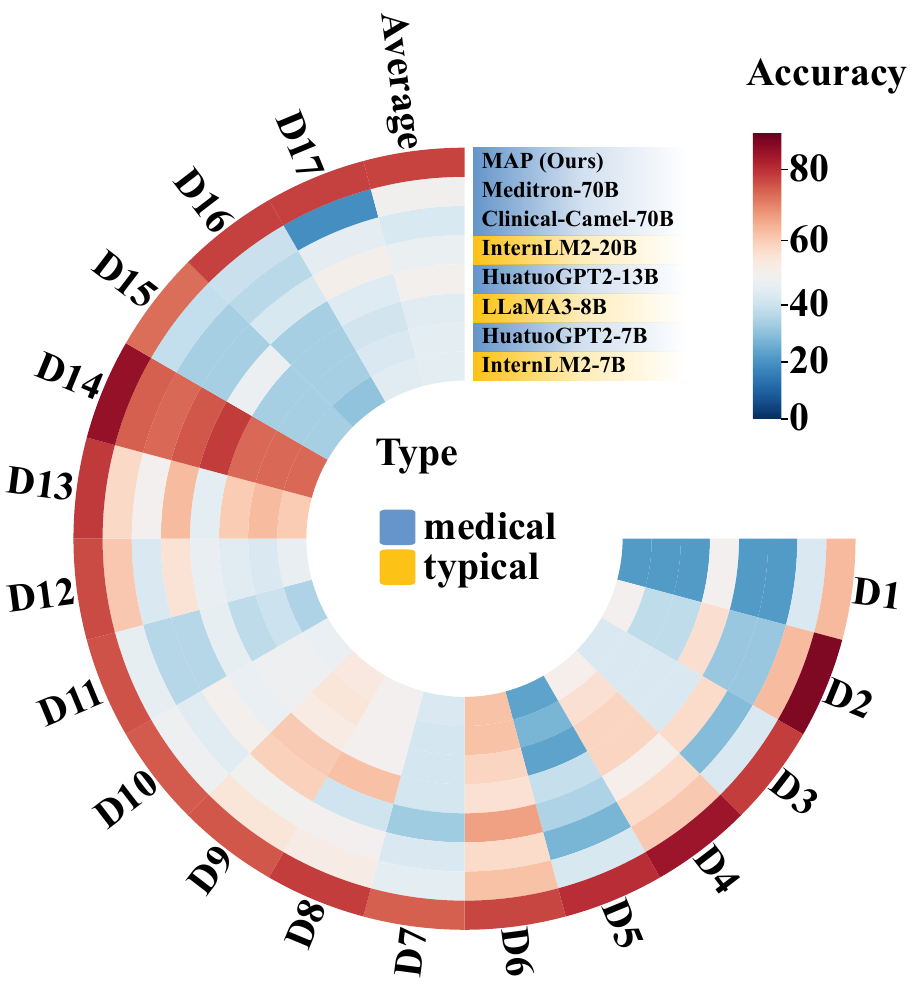}
    }
    \subfigure[Ablation Studies]{
    \includegraphics [width=0.50 \textwidth]{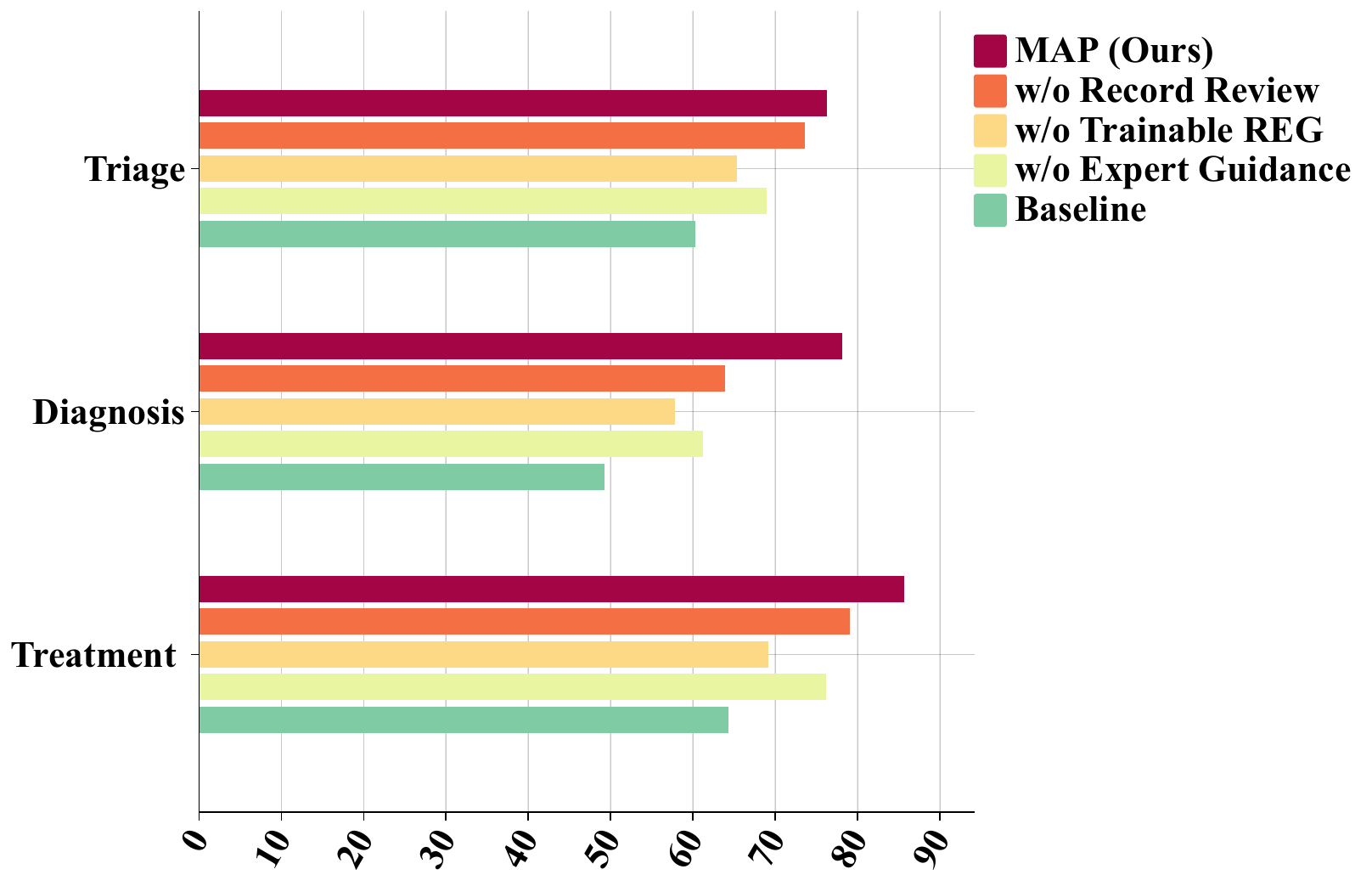}
    }
\caption{ \textbf{State-of-the-art LLMs demonstrated unsatisfying inpatient diagnostic support capabilities for complex inpatient clinical cases; in contrast, MAP enhances that capability}. (a) For instance, HuatuoGPT2-13B performed a low accuracy of 38.46\% in D5 (mental and behavioral disorders). The MAP showed significant improvement in supporting the inpatient pathways, achieving an accuracy of 79.86\% in D5. (b) Such an enhancement was attributed to the integration of our proposed record review, trainable REG, and expert guidance modules, verified by the corresponding ablation studies.}
\label{fig: as3-0}
\end{figure*}

\noindent As shown in Figure \ref{fig: as3-0} (a), state-of-the-art LLMs demonstrated notable limitations in supporting the diagnosis of complex clinical presentations involving multiple organ systems. Particularly challenging areas include D5 (mental and behavioral disorders) and D9 (diseases of the respiratory system), where the accuracy falls below 42.31\%. These findings highlighted the need for specialized diagnostic support systems designed specifically for the inpatient setting, which motivated the development of our MAP framework.

\subsection{The MAP enhances the capability of inpatient diagnostic support, particularly in complex disease cases.}
We developed the MAP, a novel multi-agent collaboration framework specifically designed to support inpatient pathways, as illustrated in Figure \ref{fig: MAP}. The MAP framework demonstrated significant improvements across all evaluation metrics, as shown in Figure \ref{fig: Dise_overall_each}. In particular, MAP achieved an overall diagnosis accuracy of 78.10\%, reflecting an 28.80\% improvement over LLaMA3-8B, which had an accuracy of 49.30\%. Notably, MAP outperformed the best specialized LLM, HuatuoGPT2-13B, by a 25.10\% improvement in accuracy (i.e., 78.10\% vs. 53.00\%).

\noindent The MAP framework exhibited significant performance in diagnosing complex conditions. For instance, as shown in Figure \ref{fig: as3-0} (a), for D5 (mental and behavioral disorders), MAP achieved an accuracy of 79.86\%, representing a 37.55\% improvement over the second-best model, Meditron-70B, which scored 42.31\%. This performance improvement can be attributed primarily to the design of the record review module, trainable REG module, and expert guidance module, whose effectiveness is verified in Figure \ref{fig: as3-0} (b). For readability, more ablation results toward more evaluation metrics were given in the Supplementary Figure \ref{fig: as2}. Benefiting from the reasoning process and diagnostic results recorded by the trainable REG module, MAP also offers explainable and clinically relevant diagnostic support.

\begin{figure*}[t]
% \flushleft
\centering
    \subfigure[Performance Comparisons]{
    \includegraphics [width=0.48 \textwidth]{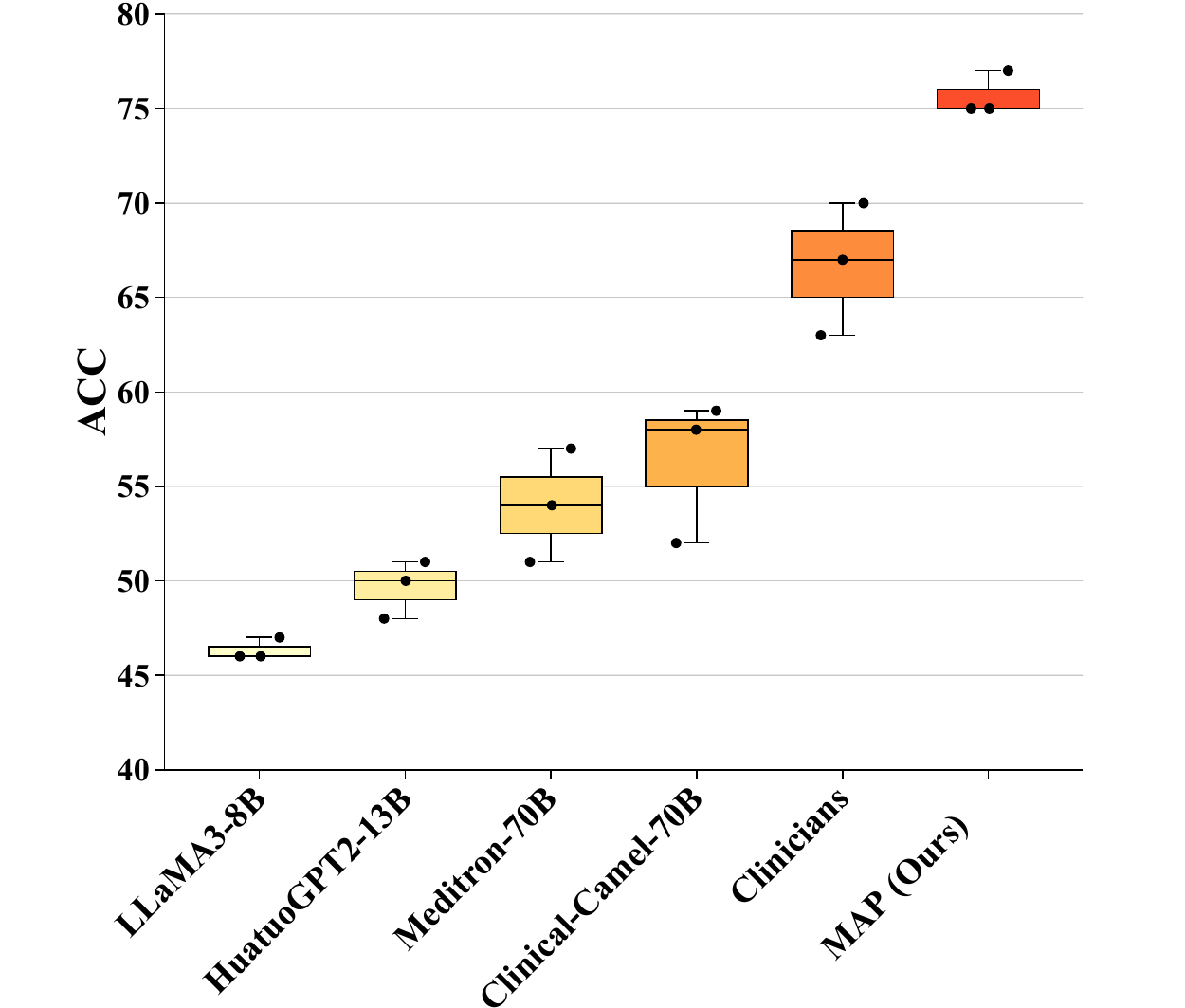}
    }
    \subfigure[Consistency Comparisons]{
    \includegraphics [width=0.45 \textwidth]{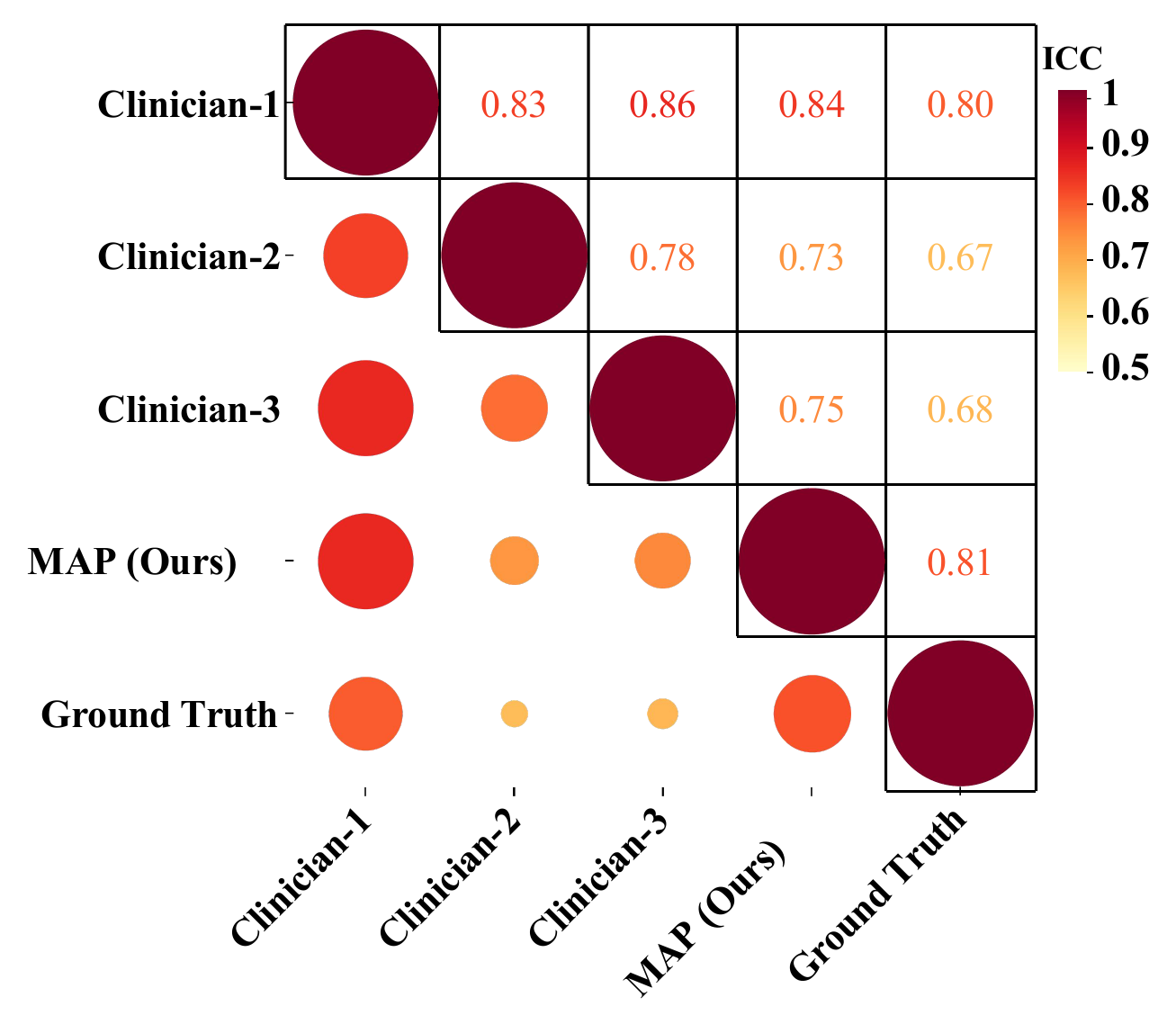}
    }
\caption{\textbf{Evaluation of LLMs performance in supporting inpatient pathways through IPDS benchmark.} (1) The MAP demonstrated consistent diagnostic support capabilities with minimal performance variance, achieving accuracy 10\%-12\% higher than board-certified clinicians (p-value\,=\,$0.0067$). (2) Statistical analysis revealed strong agreement between the MAP and the ground truth ($\textup{ICC} =0.81$), exceeding the agreement levels between individual clinicians and the ground truth ($\textup{ICC}\in[0.67,0.68]$). (3) The MAP maintained a strong alignment with clinicians($\textup{ICC}\in[0.75,0.84]$) while providing consistent diagnostic support. } 
\label{fig: clinical validation}
\end{figure*}

\subsection{Clinical validation of MAP with significant clinical compliance.}
Additionally, we conducted a statistical analysis of the diagnostic performance and decision-making abilities of clinicians by randomly selecting 100 cases from the IPDS test set to create a test set in a multiple-choice question format. Specifically, three clinicians from different tertiary hospitals were recruited to independently evaluate the patient cases, providing their reasoning and diagnostic results. Utilizing their extensive clinical experience, which ranged from 5 to 15 years, the clinicians were required to determine the appropriate triage, diagnosis, and treatment pathways for each case. Given the complexity of inpatient clinical presentations, the clinicians were permitted to provide the three most likely diagnoses and rank them according to their likelihood, facilitating the subsequent implementation of the statistical analysis. It is important to note that the primary constraints on sample size were the limited time and energy of the clinicians.

\noindent As shown in Figure \ref{fig: clinical validation} (a), the MAP outperformed the general and medical LLMs as well as clinicians, achieving the best performance statistically. In contrast, the general LLM LLaMA3-8B and the medical LLM HuatuoGPT2-13B demonstrated significantly lower performance, with accuracy levels 27\% to 30\% below those of MAP. Notably, the temperature parameter of the LLM introduces randomness into the model's classification performance. Furthermore, as illustrated in Figure \ref{fig: clinical validation} (b), the inter-rater reliability analysis revealed the degree of diagnostic correlation among three clinicians, MAP, and the ground truth. The MAP achieved a strong agreement with the ground truth, as indicated by an intra-class correlation coefficient (ICC) of $0.81$, which exceeded the agreement levels between individual clinicians and the ground truth (ICC$ \in[0.80, 0.67, 0.68]$). Additionally, the MAP maintained a strong alignment with clinician assessments (ICC$ \in[0.84, 0.73, 0.75]$), especially rivaling the diagnosis results of experienced clinicians with 15-year clinical experience, maintaining significant clinical compliance.

\section{Discussion}
Our systematic evaluation of the MAP framework in supporting inpatient pathways provided important insights into the potential and limitations of LLM-assisted inpatient pathways. We further investigated the patterns of MAP in diagnostic challenges across different disease categories, the relationship between clinical data quality and system performance, and areas for future development.

\subsection{Different diagnostic challenges across disease categories in inpatient settings.}
\begin{figure*}[t]
\centering
\includegraphics [width=0.56 \textwidth]{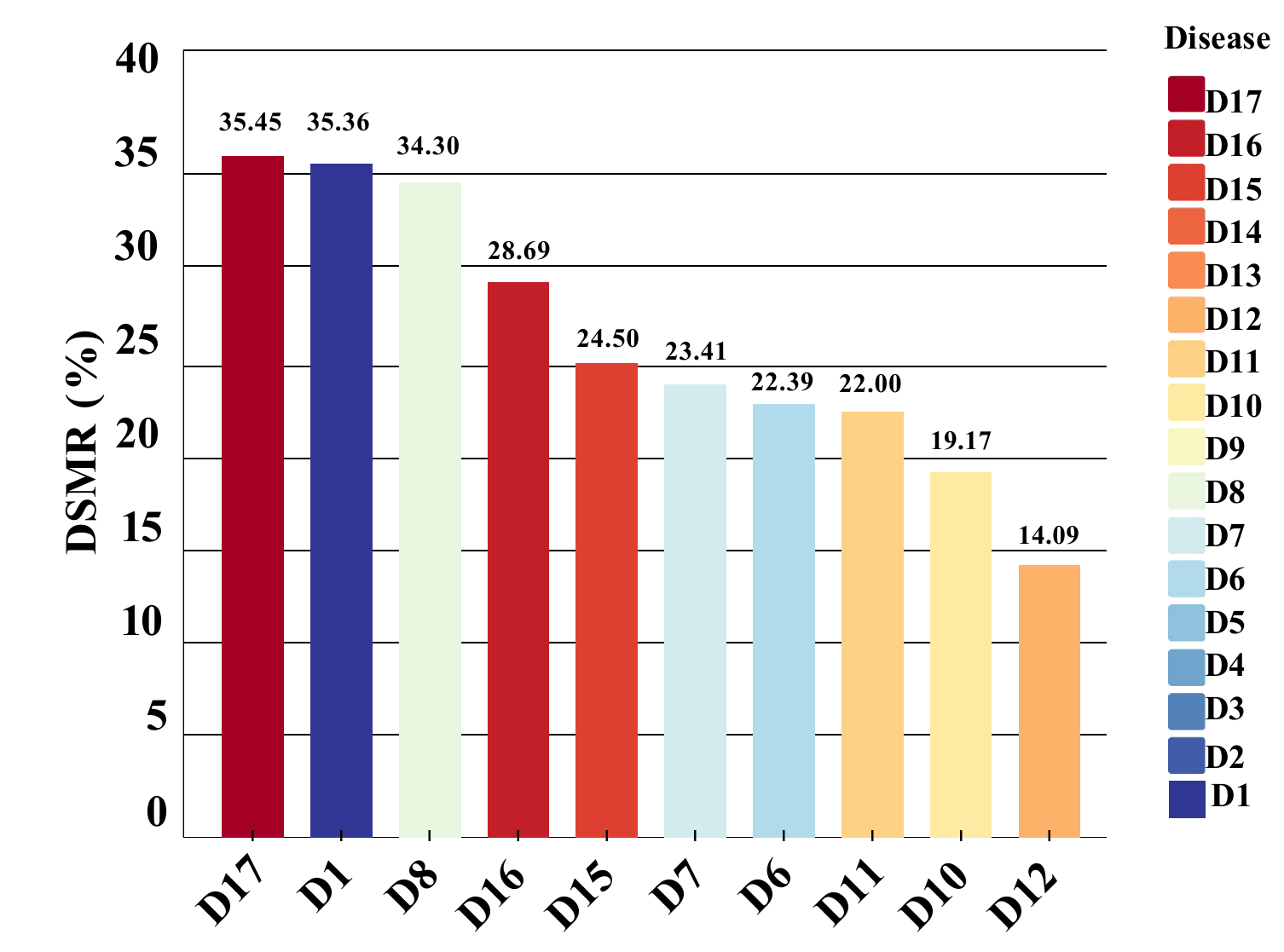}
\caption{
\textbf{Different diagnostic challenges across disease categories in inpatient settings.} We conducted the misdiagnosis rates comparison across the top 10 disease categories, where misdiagnosis rates ranged from 35.45\% to 14.09\%, demonstrating disease-specific diagnostic complexity. In particular, D1 (symptoms, signs, and abnormal clinical and laboratory findings, not elsewhere classified), D17 (certain infectious and parasitic diseases), and D8 (diseases of the skin and subcutaneous tissue) showed the highest rates (above 35.45\%), highlighting the need for improved accuracy in these areas.} 
\label{fig: Misdiagnosis}
\end{figure*}

Through systematic analysis of disease misdiagnosis rates in Figure \ref{fig: Misdiagnosis}, the study revealed significant differences in the difficulty of diagnosis across disease types. Among them, D17 (certain infectious and parasitic diseases) disease showed the highest misdiagnosis rate (35.45\%), followed by D1 (symptoms, signs, and abnormal clinical and laboratory findings, not elsewhere classified) (35.36\%), and D8 (diseases of the skin and subcutaneous tissue) (34.30\%). This prominent stratification characteristic reflects that partial diseases face unique challenges in clinical diagnosis. Of particular note is that the top three high misdiagnosis rate diseases exceeded the 34.00\% threshold, a finding highlighting the urgent need to enhance diagnostic accuracy in these specific disease areas. At the same time, the span from the highest misdiagnosis rate of 35.45\% to the lowest misdiagnosis rate of 14.09\% showed that the diagnosis complexity had apparent disease specificity.

\begin{figure*}[!htbp]
    \centering
    \subfigure[Clinician-1]{
    \includegraphics [width=0.48 \textwidth]{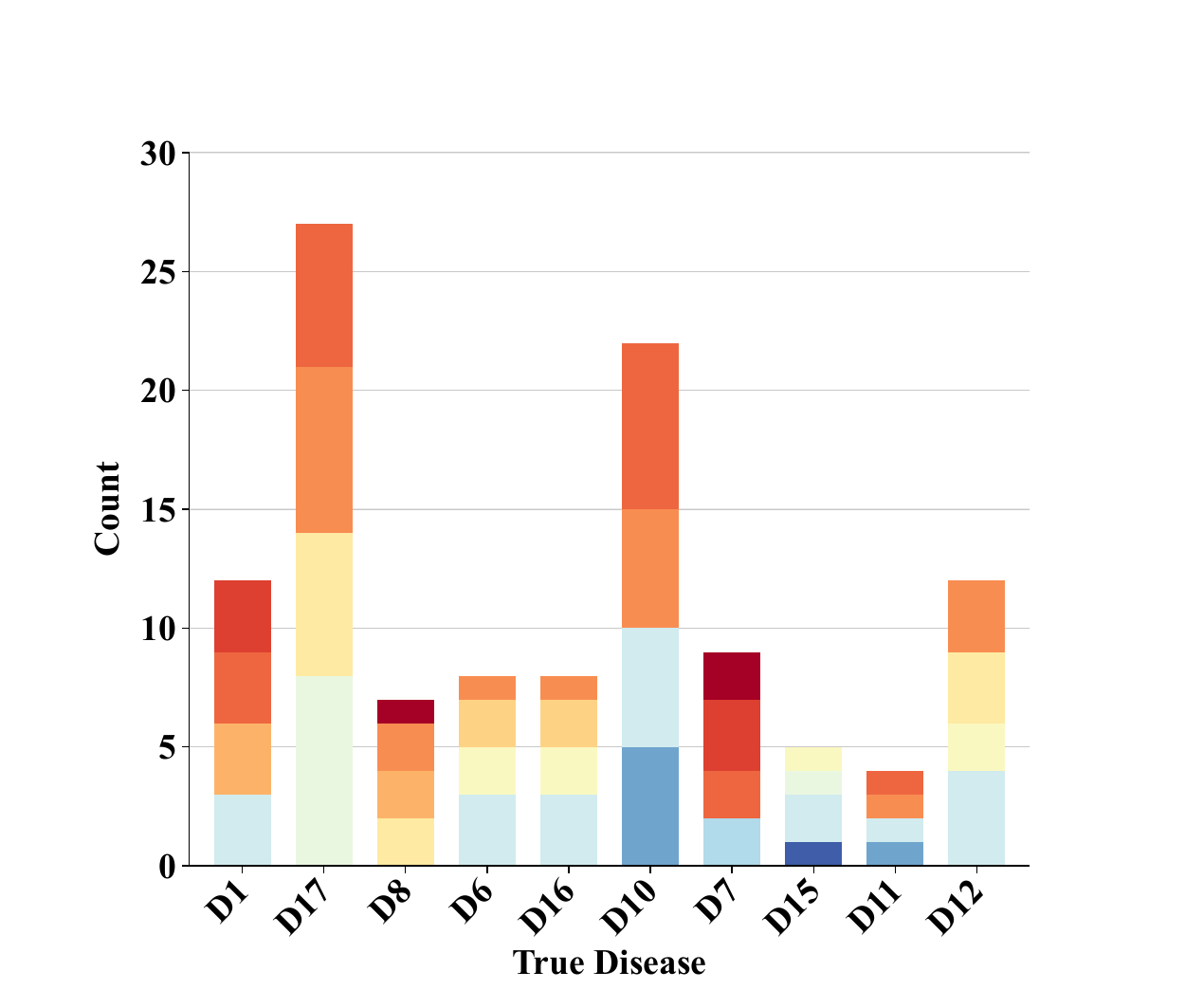}
    }
    \subfigure[MAP (Ours)]{
    \includegraphics [width=0.48 \textwidth]{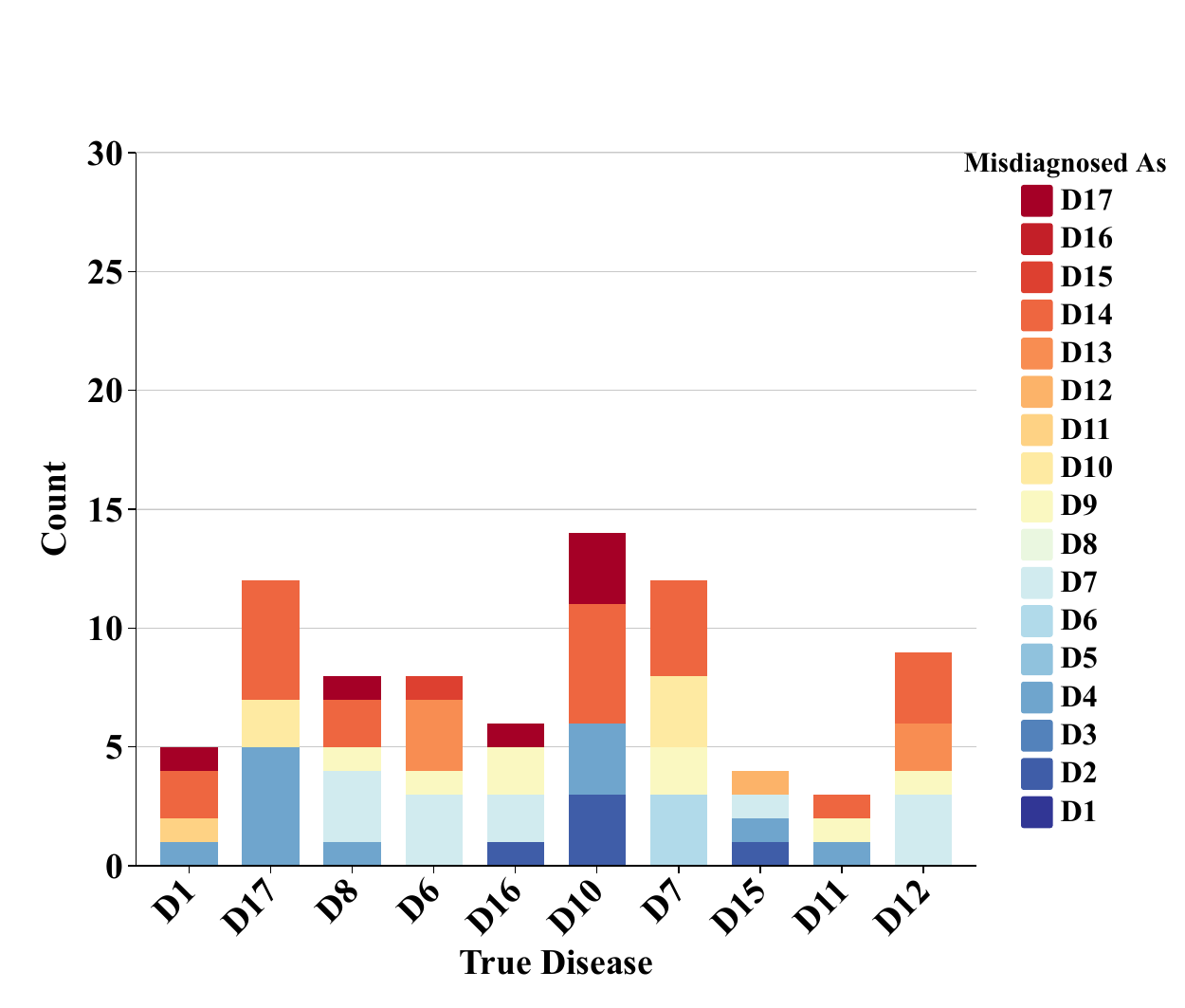}
    }
    \caption{\textbf{The MAP demonstrated superior diagnostic accuracy compared to the best-performing clinician}. In particular, the MAP significantly reduces false positives and exhibited a balanced misdiagnosis distribution across disease categories, and in contrast, the clinicians' errors were concentrated in specific categories.}
    \label{fig:conf_dise}
\end{figure*}

\subsection{The MAP showed consistently superior diagnosis across all disease categories.}
We visualized misdiagnosis patterns between Clinician-1 and the MAP toward evaluating different disease categories (D1-D17), where Figure \ref{fig:conf_dise} shows the distribution of misdiagnosis for each disease category. We found that the MAP has superior diagnostic accuracy to Clinician-1, especially in reducing false positive samples and maintaining diagnostic specificity across different disease categories. Moreover, the MAP showed the most balanced misdiagnosis distribution, with consistently high diagnosis counts across all disease categories. The best-performance Clinician-1 misdiagnoses were more pronounced in specific disease categories D1 (symptoms, signs, and abnormal clinical and laboratory findings, not elsewhere classified) and D17 (certain infectious and parasitic diseases) but with lower overall misdiagnosis counts, showing moderate performance with concentrated but significant misdiagnosis patterns.

\begin{figure*}[!]
\centering
    % \includegraphics [width=0.98 \textwidth]{Figures/Figure 8_v3.pdf}
    % \subfigure[Different Clinical Data Input]{
    % \includegraphics [width=0.43 \textwidth]{Figures/as_acc_2.pdf}
    % }
    % \subfigure[Specific Disease Analyses]{
    \includegraphics [width=0.8 \textwidth]{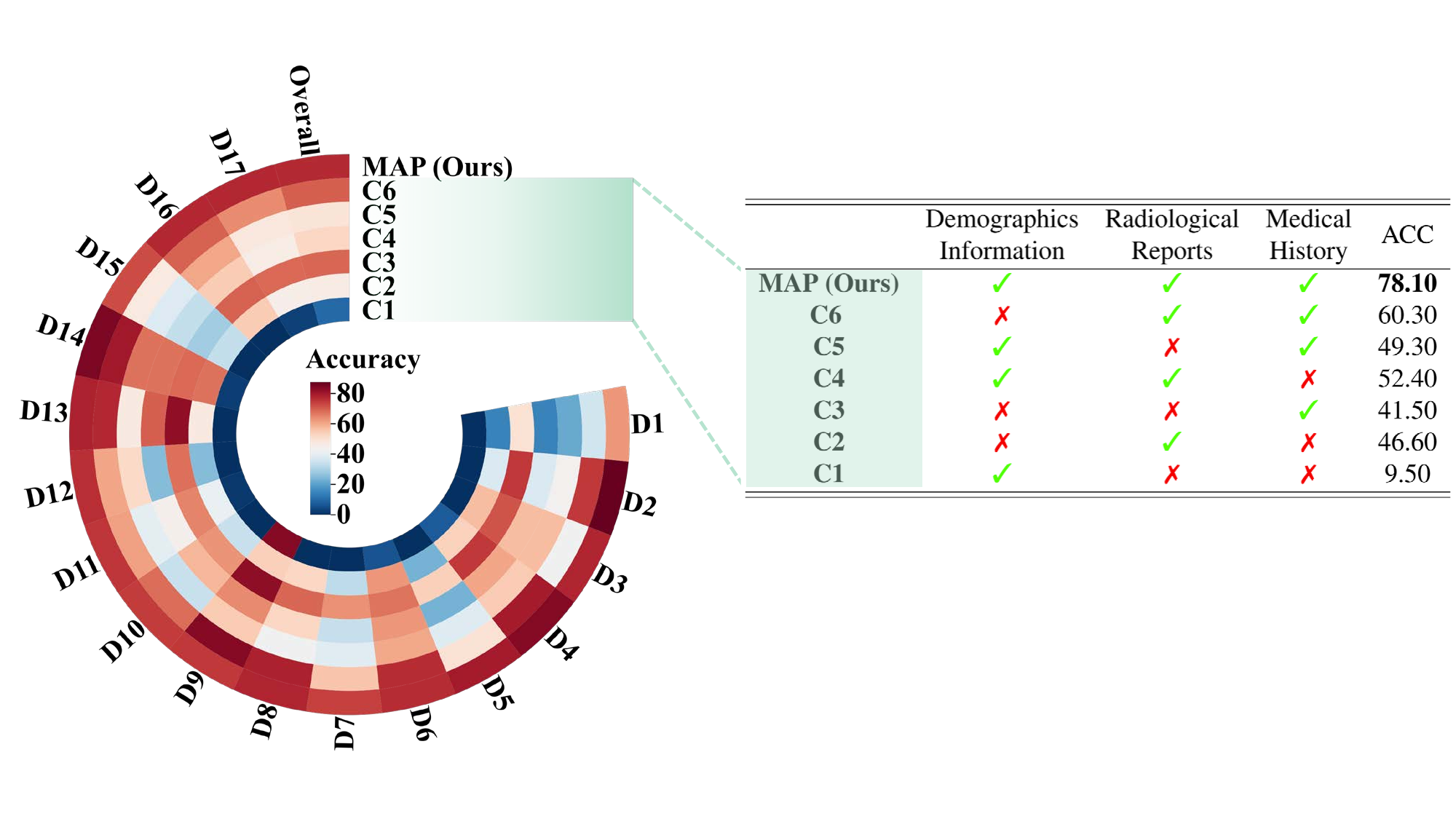}
    % }
    \caption{\textbf{Impact of clinical data input on inpatient diagnostic support performance.} (1) Comparison of comprehensive medical history improves diagnostic support accuracy from 23.08\% to 79.86\% for D5 (mental and behavioral disorders); (2) Incorporation of radiological findings enhanced diagnostic support accuracy by 31.03\% (from 47.50\% to 78.53\%) for D13 (diseases of the digestive system); (3) Demographic information contributed a 5.80\% improvement in overall diagnostic support accuracy (from 46.60\% to 52.40\%). These findings highlighted the importance of systematic integration of clinical data components, particularly emphasizing the value of medical history and radiological data in supporting inpatient diagnostic decisions.} 
\label{fig: as_input}
\end{figure*}

\subsection{Impact of input data on inpatient pathways.}
We found that including or excluding specific input data could substantially affect the performance of LLMs, evidenced by ablation experiments of input components such as demographic information, radiological reports, and medical history on the IPDS benchmark toward inpatient pathways, as shown in Figure \ref{fig: as_input}. We observed that: ($i$) These input components played a critical role in integrating patients' medical history for accurate diagnosis, particularly in complex, chronic conditions. The significant 55.02\% increase in the MAP's diagnostic accuracy, from 23.08\% to 78.10\%, for patients with D5 (mental and behavioral disorders), when medical history was included, is a reassuring validation of the value of our work since historical data provided essential context for interpreting current symptoms. ($ii$) Radiological reports also play a pivotal role in enhancing diagnostic accuracy, especially for internal conditions. The 31.03\% improvement from 47.50\% to 78.53\% in the diagnosis of D13 (diseases of the digestive system) when radiological report data was included. This underscores the importance of integrating imaging data into LLM-assisted inpatient pathways. ($iii$) Our research found that while demographic information contributed to the diagnosis task, its impact was less pronounced than medical history and radiological report data. Including demographic information improved overall accuracy by only 5.80\% (from 46.60\% to 52.40\%), suggesting that while relevant, this information plays a supporting role in inpatient pathways. These findings underscore the urgent need for comprehensive and diverse data inputs in LLM-assisted inpatient pathways. They also prioritize certain data types in time-critical situations, particularly medical history, and radiological reports, emphasizing the importance of these data types in enhancing diagnostic accuracy and the critical role of our research in addressing this need.

\subsection{Limitations and future directions.}

Our findings underscored the potential of our approach in significantly improving LLMs in inpatient pathways, inspiring the audience with the possibilities of AI in healthcare. Despite these impressive results, our study also identified areas requiring further improvement. First, it was expected to introduce out-patient data \cite{wan2024outpatient,mani2024causes} to improve LLMs in both inpatient and out-patient diagnosis significantly, consequently enriching the benchmark and further developing the emergency medicine community. Furthermore, improving the decision-making explainability of the MAP emerged as a vital area for future development to enhance trust and adoption among medical professionals, especially in complex cases where the reasoning process was not immediately apparent to clinicians. Moreover, it was crucial to emphasize that the MAP was designed to augment, not replace, the expertise of medical professionals. The collaborative model of LLMs and human clinical judgment remained essential for optimal patient care in inpatient settings. In conclusion, our results provided a foundation for future advancements in LLMs, ultimately aimed to improve patient outcomes within real-world inpatient settings.

\section{Online content}
Any methods, additional references, Nature Portfolio reporting summaries, source data, extended data, supplementary information, acknowledgments, peer review information, details of author contributions and competing interests, and statements of data and code availability are available at the link.

\section{References}
\begin{spacing}{0.9}
\bibliographystyle{naturemag}
\bibliography{icu_agents}
\end{spacing}

\newpage

\section{Benchmark preparation}
We curated the IPDS, an inpatient diagnostic support dataset integrating three information modules from the Medical Information Mart for Intensive Care (MIMIC) database, namely MIMIC-IV-Hosp, MIMIC-IV-ICU, and MIMIC-IV-Note. Specifically, the MIMIC-IV-Hosp module, containing data on 546,028 unique hospitalizations for 223,452 individuals, provided demographic information (language, marital status, race, gender), diagnosis information (ICD code, ICD version, long title, disease) \cite{lovaasen2012icd}, and treatment pathway options (current service).
The MIMIC-IV-ICU module, with data on 94,458 ICU stays for 65,366 unique individuals, contributed hospital admission information, specifically the department distribution.
The MIMIC-IV-note module included 331,794 de-identified discharge summaries and 2,321,355 de-identified radiological reports, where we extracted the medical history and radiological reports of patients. We began with the medical information consolidation by identifying missing values, e.g., samples with null radiological report values were filtered out. Next, we conducted the attribute selection identification to select the most relevant diagnostic features, including demographic information, radiological reports, medical history, etc. Finally, we achieved the comprehensive data screening to delete some distractors, e.g., the medical history data was eliminated since it was utterly unrelated to the radiological reports in some patient cases. The specific abbreviations and details of the department, disease, and treatment options were detailed in Table \ref{tab: depa_table}, \ref{tab: dise_table}, and \ref{tab: trea_table}, respectively.

\noindent Using unique subject identifiers, we performed multi-table search and integration operations to construct the IPDS benchmark, comprising 51,274 patient cases. Each case incorporates curated demographic information, radiological reports, and medical history. Remarkably, for \textbf{the triage task}, IPDS selected the clinic category column data assigned at the time of admission as the patient's triage category, involving nine departments. For \textbf{the diagnosis task}, IPDS reclassified the original 1,298 disease labels into 17 broader categories using an international statistical disease classification system. To prioritize clinical relevance, IPDS selected the three highest-priority diagnostic indicators from MIMIC-IV, where the highest-priority label was typically designated as the primary diagnosis. Each patient case includes comprehensive diagnostic metadata, such as ICD codes (ICD-9/10), standardized disease titles, categorical labels, and priority indicators, ensuring clinically accurate representation. For \textbf{the treatment task}, IPDS selected the treatment plan column data mentioned in the MIMIC-IV service table as the patient's designated treatment category, involving 16 options. This multifaceted structure enables the evaluation of state-of-the-art LLMs across critical stages of clinical decision-making (e.g., diagnostic prioritization) within the context of inpatient pathways.

\begin{table*}[t]
\centering
\caption{\textbf{Triage options in the IPDS benchmark.}}  
\label{tab: depa_table}
\resizebox{0.6\linewidth}{!}{
\begin{tabular}{l|l}
\hline\hline
Abbreviation       & Details                                   \\
\hline
CVICU              & Cardiac Vascular Intensive Care Unit     \\
CCU                & Coronary Care Unit                       \\
MICU               & Medical Intensive Care Unit              \\
MSICU              & Medical and Surgical Intensive Care Unit \\
Neuro Intermediate & Neuro Intermediate                       \\
Neuro Stepdown     & Neuro Stepdown                           \\
Neuro SICU         & Neuro Surgical Intensive Care Unit       \\
SICU               & Surgical Intensive Care Unit             \\
TSICU              & Trauma Surgical Intensive Care Unit     \\
\hline\hline
\end{tabular}
}
\end{table*}

\begin{table*}[t]
\caption{\textbf{Diagnosis options in the IPDS benchmark.}}  
\label{tab: dise_table}
\resizebox{0.98\linewidth}{!}{
\begin{tabular}{l|l}
\hline\hline
Abbreviation & Details                                                                                                \\
\hline
D1    & Symptoms, signs, and abnormal clinical and laboratory findings, not   elsewhere classified             \\
D2    & Symptoms, signs and abnormal clinical and laboratory findings                                         \\
D3    & Pregnancy, childbirth and the puerperium                                                              \\
D4    & Neoplasms                                                                                             \\
D5    & Mental and behavioral disorders                                                                      \\
D6    & Injury, poisoning, and certain other consequences of external causes                                   \\
D7    & Endocrine, nutritional and metabolic diseases                                                         \\
D8    & Diseases of the skin and subcutaneous tissue                                                          \\
D9    & Diseases of the respiratory system                                                                    \\
D10   & Diseases of the nervous system and sense organs                                                       \\
D11   & Diseases of the musculoskeletal system and connective tissue                                          \\
D12   & Diseases of the genitourinary system                                                                  \\
D13   & Diseases of the digestive system                                                                      \\
D14   & Diseases of the circulatory system                                                                    \\
D15   & Diseases of the blood and blood-forming organs and certain disorders   involving the immune mechanism \\
D16   & Congenital malformations, deformations and chromosomal abnormalities                                  \\
D17   & Certain infectious and parasitic diseases \\
\hline\hline
\end{tabular}
}
\end{table*}

\noindent The de-identification process was rigorous and multi-faceted, where the IPDS used random ciphers to replace patient identifiers and applied stringent rules to structured columns. Text fields were filtered using manually curated allow and block lists, and the IPDS implemented context-specific regular expressions. A free-text de-identification algorithm was applied, followed by a manual review to ensure the complete removal of personally identifiable information. This thorough process ensured patient privacy while maintaining the clinical compliance and integrity of the data. The dataset was distributed as comma-separated value files, with each row representing a unique case identified by a unique subject\_id and hadm\_id. Access to the IPDS required registration on PhysioNet, identity verification, completion of human participant training, and signing of a data use agreement. These measures ensured the ethical use of the data while maintaining its research value. To facilitate research and collaboration, we established an open-source MIMIC Code Repository, serving as a platform for shared discussion and analysis of all versions of MIMIC, including the IPDS.

\begin{table*}[!ht]
\caption{\textbf{Treatment options in the IPDS benchmark.}}  
\label{tab: trea_table}
\resizebox{0.98\linewidth}{!}{
\begin{tabular}{l|l}
\hline\hline
Index & Detailed Text                                                                                           \\
\hline
T1    & Vascular surgery, mainly refers to surgery related to the circulatory system                         \\
T2    & Thoracic surgery, mainly refers to chest surgery between the abdomen and the neck                    \\
T3    & Trauma surgical treatment, physical injury or damage caused by external physical factors             \\
T4    & General surgical treatment, mainly refers to types of surgery that cannot be classified by specialty \\
T5    & Plastic treatment, mainly for the repair or reconstruction of the human body                         \\
T6    & Orthopedic surgical treatment, mainly involving the musculoskeletal system                           \\
T7    & Orthopedic treatment, mainly involving the musculoskeletal system                                    \\
T8    & Obstetrics, maternal classification and refusal                                                      \\
T9    & Neurosurgery treatment, surgical treatment related to the brain                                      \\
T10   & Neurology treatment, non-surgical treatment related to the brain                                     \\
T11   & General medical treatment                                                                              \\
T12   & Gynecological treatment, female reproductive system and breasts, etc.                                \\
T13   & Urogenital treatment, urinary system and reproductive system                                         \\
T14   & Otolaryngology treatment mainly for the ear, nose and throat-related areas                           \\
T15   & Cardiovascular surgery treatment, surgical treatment of cardiovascular diseases                      \\
T16   & Cardiovascular medicine treatment, conservative treatment of cardiovascular diseases\\
\hline\hline
\end{tabular}
}
\end{table*}

% 这里是evaluation部分
% The multi-agent characteristic is reflected in the setting that 
\section{Methods}

The Multi-Agent Inpatient Pathways (MAP) framework, illustrated in Figure \ref{fig: MAP}, implements a Triage-Diagnosis-Treatment (TDT) clinical pathway through collaboration among three clinician agents and a chief agent. In the MAP framework, each agent is empowered by a specialized LLM augmented with domain-specific medical knowledge (e.g., ICD-10 codes, NICE guidelines), and is capable of processing and understanding complex medical scenarios. We organize the communication among agents using a structured protocol with three fields, including the context, thinking, and answer. The context field contains a brief summary of the current task and the conversation history among agents. The thinking field is composed of the reasoning of the agent and provides transparency to help other agents understand better. The answer field is the decision of the agent in the format of the concise description of standardized medical terminology with ICD-10 codes. The structured protocol makes the interaction of agents well-informed and efficient, which is traceable and interpretable in clinical deployment.

\noindent Along the inpatient pathway, the first Triage Agent prioritizes patient urgency by comprehensively analyzing radiology reports (e.g., Modified Early Warning System) and medical history (e.g., Hypertension, Hyperlipidemia, Atrial Fibrillation/flutter), and assigns patients to one of nine departments (e.g., Coronary Care Unit). Then, the Diagnosis Agent, deployed within the specific department, takes the output of the first Triage Agent and the demographic information, radiological reports, medical history, routine examination, and preliminary evaluation results as input and employs chain-of-thought reasoning to generate accurate diagnoses. Afterwards, the Treatment Agent continued to determine the appropriate treatment plan for the patient based on the analysis provided by the preceding two clinical agents and the patient's information, and completed the entire inpatient pathway process. These three clinician agents are enhanced with our tailored modules, including the record review module, the trainable retrieval-enhanced generation module, and the expert guidance module. In particular, the Chief Agent, involved in the expert guidance module during the training, took the patient's condition and was responsible for overseeing these clinician agents, judging whether their predictions of the triage, diagnosis and treatment were convincing or not. The Chief Agent would point out the mistakes and improper reasonings effectively, and provide structured guidance to remind the clinician agents to reconsider and learn from such supervision.

\noindent The record review module is designed to select the beneficial input data (e.g., the medical history records and radiological reports), aiming to address cases where the medical history records contradict the radiological reports. This procedure ensures that only relevant and meaningful data are considered in the diagnostic process. Specifically, we used ClinicalBERT \cite{huang2019clinicalbert} to embed the medical entities and the overall report into a correlation coefficient matrix, where each element represents the degree of correlation between a specific medical fact in the medical history and the features in the radiology report. Then, we applied the cosine similarity function to calculate the importance score of each entity. The key entities with high scores are used as input for correlation analysis. For example, an inpatient case with a high correlation score is calculated when the radiology report emphasizes lung abnormalities and the medical history mentions \textit{pulmonary nodules}. We set a threshold for the correlation coefficient (empirically set as 0.1), and considered medical history with a correlation coefficient lower than the correlation threshold to be irrelevant to the radiology report, thus only considering reliable information from the radiology report for the clinical decision. Therefore, the record review module ensures that irrelevant or noisy medical history data do not interfere with diagnostic reasoning with the on-site radiological report, improving the quality of input data for diagnosis and reducing the risk of errors caused by unrelated information.

\noindent The trainable retrieval-enhanced generation module first conducted the data retrieval for each patient case, including demographic information, radiological reports, and medical history. Then, it integrated a comprehensive collection of real patient cases from clinical practice alongside authoritative medical guidelines from the National Institute for Health and Care Excellence (NICE) \cite{NICE} professional medical databases to generate the knowledge base. Using the vector store architecture of LlamaIndex \cite{Liu_LlamaIndex_2022}, all documents in the knowledge base are embedded to produce dense vector representations that effectively capture clinical semantic relationships. Afterward, the system leveraged the semantic search capabilities of LlamaIndex to retrieve the top 10 most relevant documents from the knowledge base using cosine similarity. To further enhance the model's diagnostic reasoning capabilities, a structured Chain-of-Thought (CoT) reasoning tool was incorporated to include the diagnostic reasoning process, its derived ultimate diagnosis, and supporting evidence in the input data. This context enhancement ensures that the training data incorporates highly relevant similar cases, medical guidelines, and reasoning processes for each instance, improving the interpretability and robustness of its diagnostic conclusions.

\noindent The expert guidance module was designed to provide guidance to facilitate the reasonable diagnostic process performed by the diagnosis agent. Specifically, a chief agent was designed to systematically evaluate the diagnosis agent's diagnostic conclusions against established medical standards and guidelines of the knowledge base, i.e., it verified whether the proposed diagnosis was supported by sufficient evidence, aligned with clinical guidelines, and adhered to standard diagnostic procedures. When inconsistencies in the diagnostic reasoning were detected, the chief agent pinpointed the specific issues. For example, if the diagnosis agent overlooked the crucial lung nodule feature in the radiological report, the chief agent may generate guidance: \textit{The lung nodule feature mentioned in the radiological report seems relevant to this case but was not considered in the diagnostic reasoning.} In summary, the expert guidance module introduced guidance and reflection on the reasoning process through the knowledge base, enhancing the diagnostic reasoning capability of the model.

\noindent In the implementation, all patient data were de-identified, ensuring privacy while preserving essential clinical fields, including basic patient information, medical history, and radiological findings. The primary objective of training the model is to teach it to effectively filter out retrieved but irrelevant cases that do not contribute to diagnosing the current patient. This reduces potential interference from semantically similar but clinically irrelevant data. We selected LLaMA-3-8B as the base model for four specialized agents in our MAP framework. Three clinician agents equipped with three tailored modules were optimized with 3 epochs with a learning rate of $2 \times 10^{-5}$, and the chief agent was equipped with a standard retrieval-augmented generation with real patient cases from clinical practice alongside authoritative medical guidelines to provide high-quality guidance. During the training, we froze the LLM parameters while fine-tuning the learnable parameters in the trainable retrieval-enhanced generation module to avoid overfitting. During the inference, three clinician agents in the MAP framework are evaluated with the record review module and the trainable retrieval-enhanced generation module, since the expert guidance module is designed to enhance the training process. All experiments were performed on a Linux platform with eight NVIDIA A800 GPUs.

\begin{table*}[t]
\centering
\caption{\textbf{Data source of the IPDS benchmark}}
\label{tab: data_source}
% \resizebox{0.88\linewidth}{!}{ 
\begin{tabular}{lll}
\hline\hline
\textbf{Input}                       & \textbf{Source}                     & \textbf{Detailed File}                     \\
\hline\hline
language, marital status, race       & MIMIC-IV-Hosp                       & admissions.csv                             \\
gender                               & MIMIC-IV-Hosp                       & patients.csv                               \\
icd\_code, icd\_version, long\_title, disease & MIMIC-IV-Hosp                       & d\_icd\_diagnoses.csv             \\
curr\_service                         & MIMIC-IV-Hosp                       & services.csv                              \\
medical history                 & MIMIC-IV-Note                       & discharge.csv                              \\
radiological report                          & MIMIC-IV-Note                       & radiology.csv                              \\
department                           & MIMIC-IV-ICU                        & icustays.csv                               \\
% disease              & MIMIC-IV-Hosp                       & d\_icd\_diagnoses.csv                                       \\
                                     % & List of ICD-9/10 codes              & Wikipedia                                  \\
\hline\hline
\end{tabular}
\end{table*}

\section{Evaluation metrics}

\noindent To comprehensively evaluate the performance of LLMs in the inpatient setting, we employed a series of evaluation metrics, including accuracy, macro precision, macro F1-score \cite{powers2020evaluation}, sensitivity \cite{yerushalmy1947statistical}, specificity \cite{saah1998sensitivity}, Cohen's Kappa (CK) \cite{mchugh2012interrater}, Matthews Correlation Coefficient (MCC) \cite{matthews1975comparison}, and Area Under the receiver operating characteristic Curve (AUC) \cite{lobo2008auc}. Moreover, to meet the needs of clinical diagnosis, we utilized the Disease-Specific Misdiagnosis Rate (DSMR), which quantified the proportion of misdiagnosed cases (false positives) relative to the total number of actual cases for a specific disease. It could be mathematically expressed as $\text{DSMR} = n\_misd/\ n\_total \times 100\%$, where $n\_misd$ denoted the number of misdiagnoses with both false positives and false negatives, $n\_total$ denoted the number of diagnoses misdiagnoses. We also applied the two-way random effects Intraclass Correlation Coefficient $\text{ICC}(2,k)$\cite{koo2016guideline} to analyze the rating reliability since it is well-suited to that case multiple raters independently evaluate the same subjects. It could be mathematically expressed as $(\text{MSR} - \text{MSE})/(\text{MSR} + (1/n)\times (k \cdot \text{MSC} - \text{MSE}))$, where MSR, MSE, MSC, $n$, and $k$ denote the Mean Square for Rows, Mean Square Error, Mean Square for Columns, the number of subjects, and the number of raters, respectively. The ICC reference table is as follows: $\text{ICC}<0.40: \text{Poor reliability}; 0.40\leq\text{ICC}<0.60: \text{Fair reliability}; 0.60\leq \text{ICC}< 0.75: \text{Good reliability}; \text{ICC}\geq0.75:\text{Excellent reliability}$.

\section{Ablation study and qualitative case study}
\begin{figure*}[h!]
\centering

    \subfigure[Triage]{
   \includegraphics [width=0.628 \textwidth]{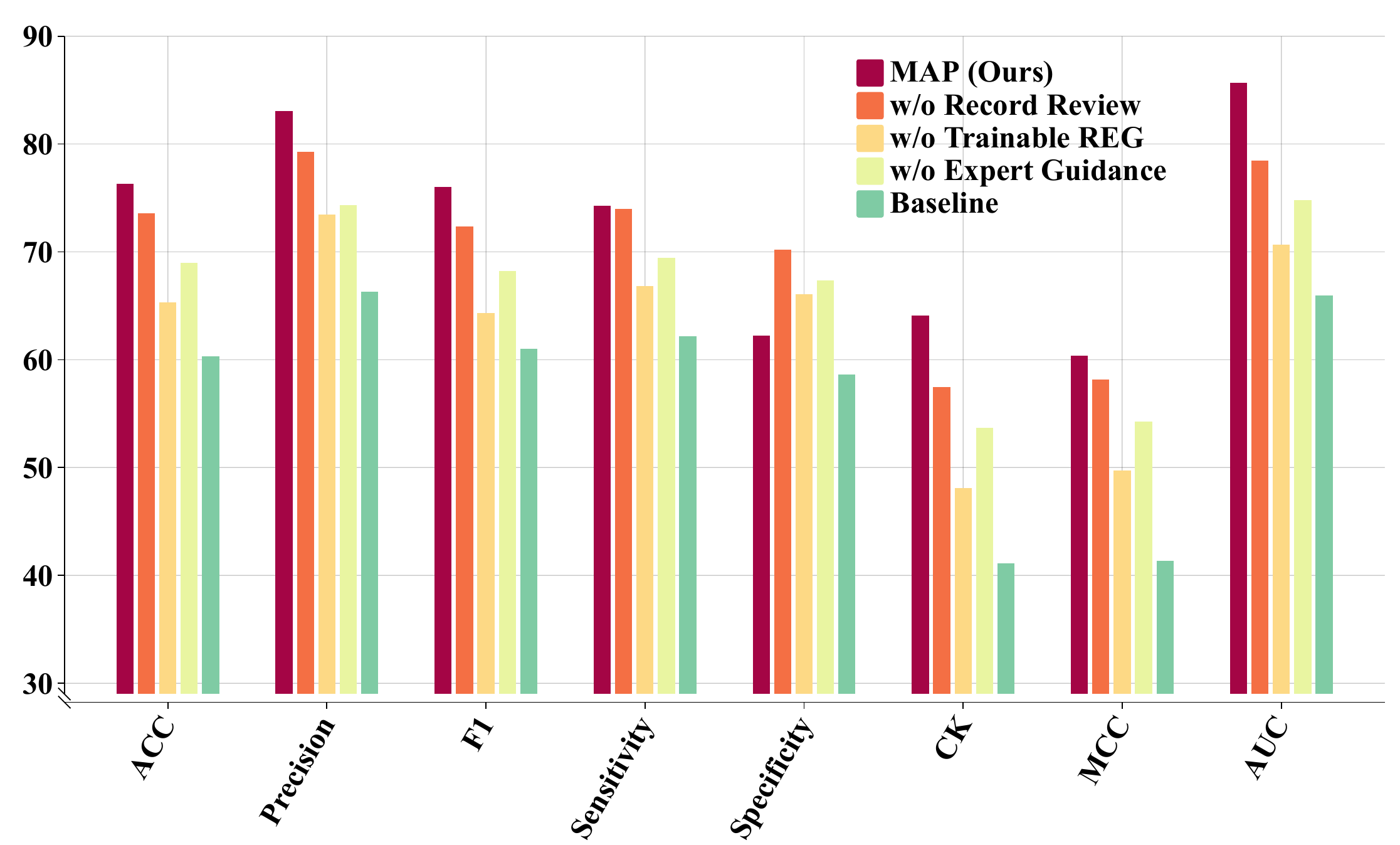}
   }
   \subfigure[Diagnosis]{
   \includegraphics [width=0.628 \textwidth]{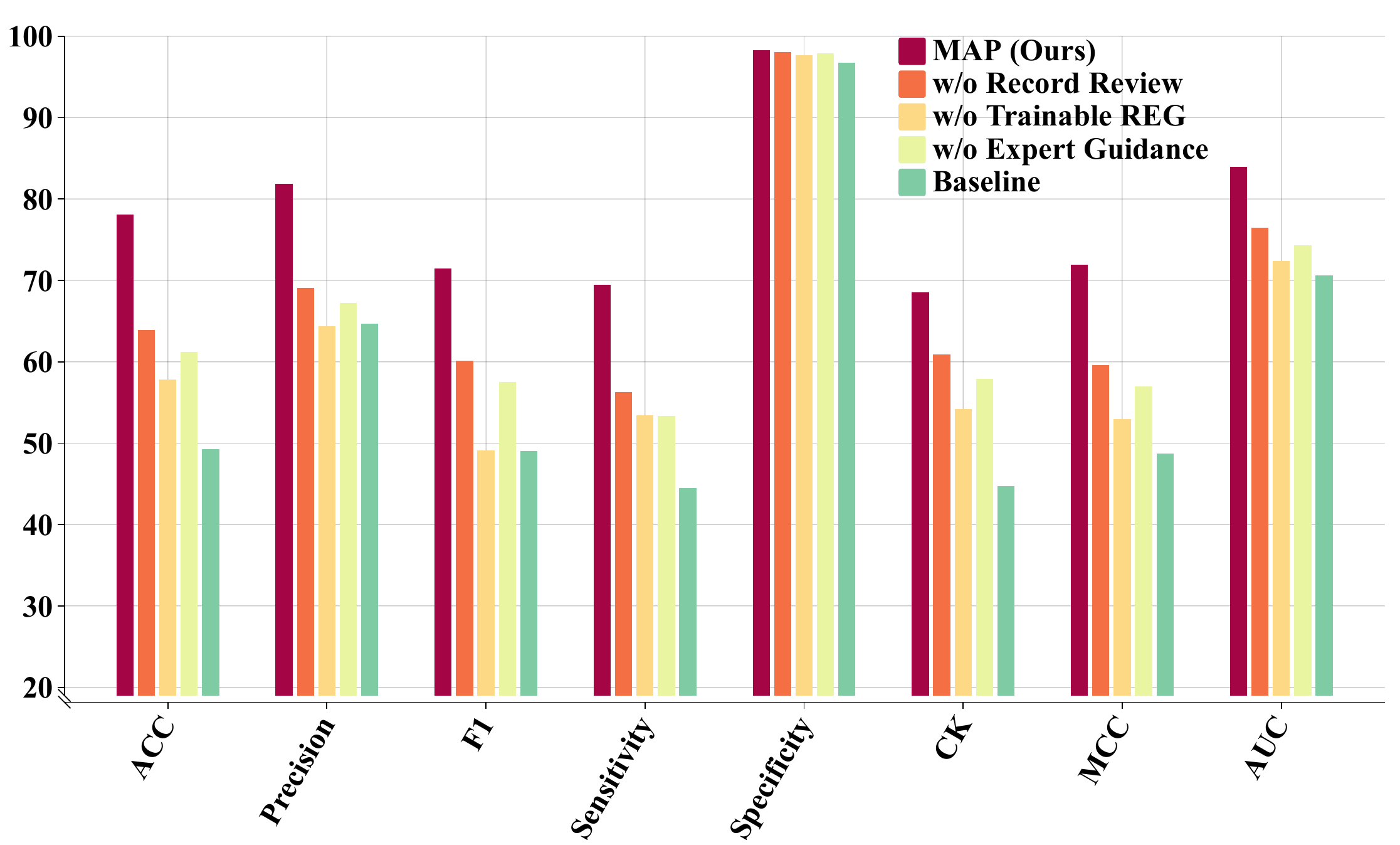}
   }
    \subfigure[Treatment]{
   \includegraphics [width=0.628 \textwidth]{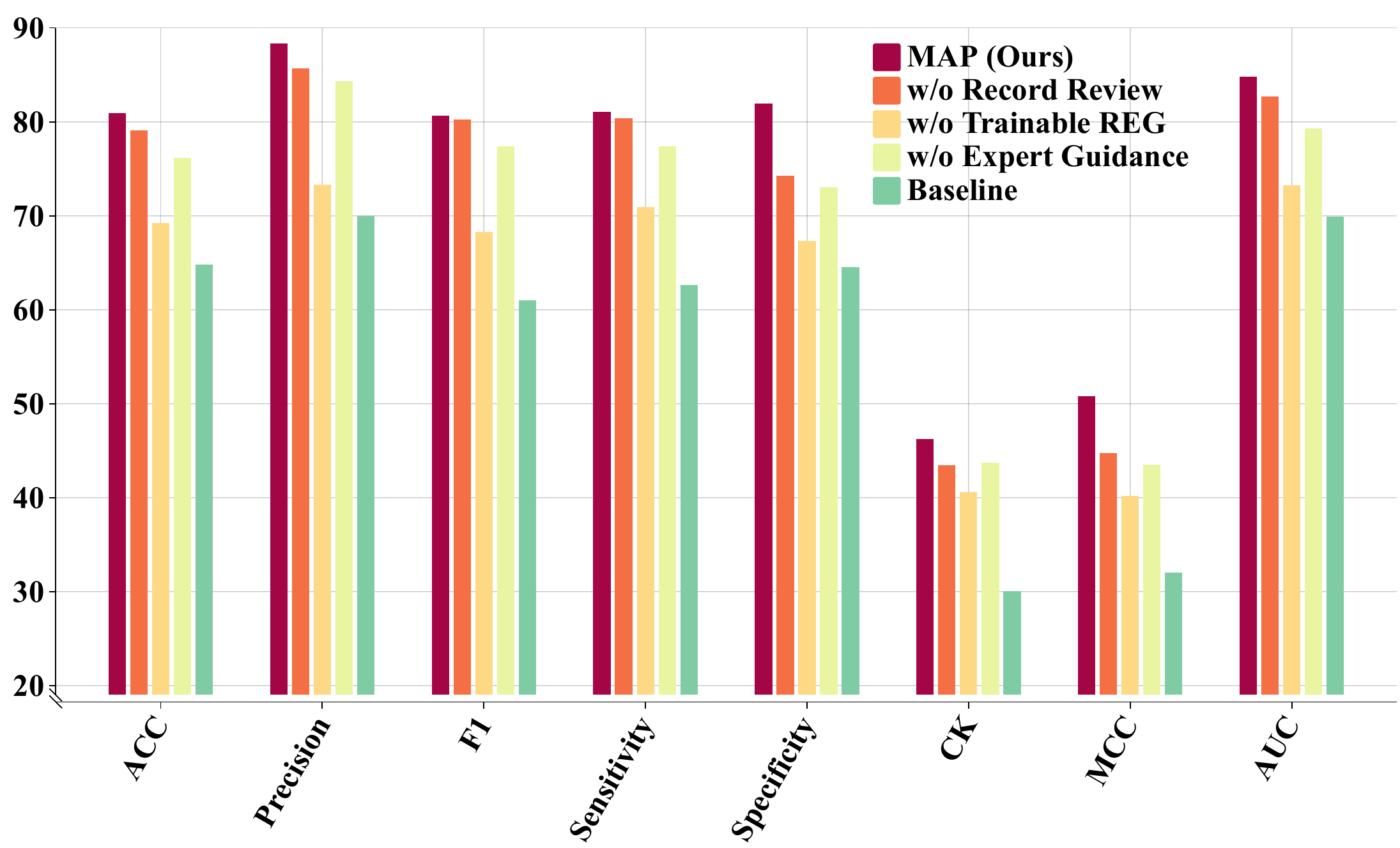}
   }
\caption{\textbf{Comprehensive ablation studies of the record review, trainable REG, and expert guidance modules} on inpatient pathway tasks with eight evaluation metrics.}
\label{fig: as2}
\end{figure*}

\begin{figure*}[h]
\centering
   \includegraphics [width=0.88\textwidth]{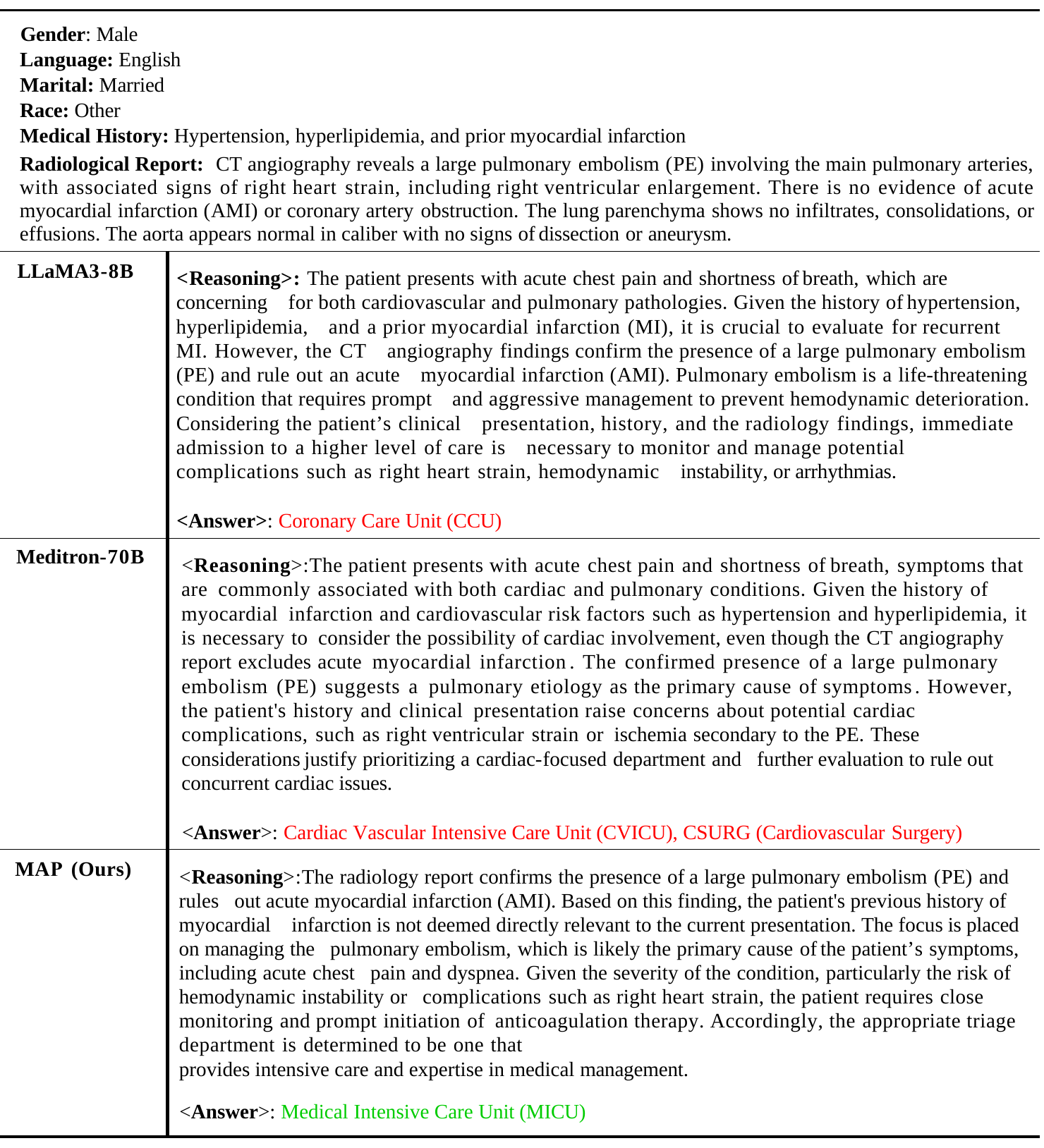}
\caption{The comparison shows the triage and its reasoning of LLaMA3-8B, Meditron-70B, and our MAP. For better visualization, we highlight the incorrect parts in \textcolor{red}{red} and the correct parts in \textcolor{green}{green}.}
\label{fig: example2}
\end{figure*}

\begin{figure*}[h]
\centering
   \includegraphics [width=0.88\textwidth]{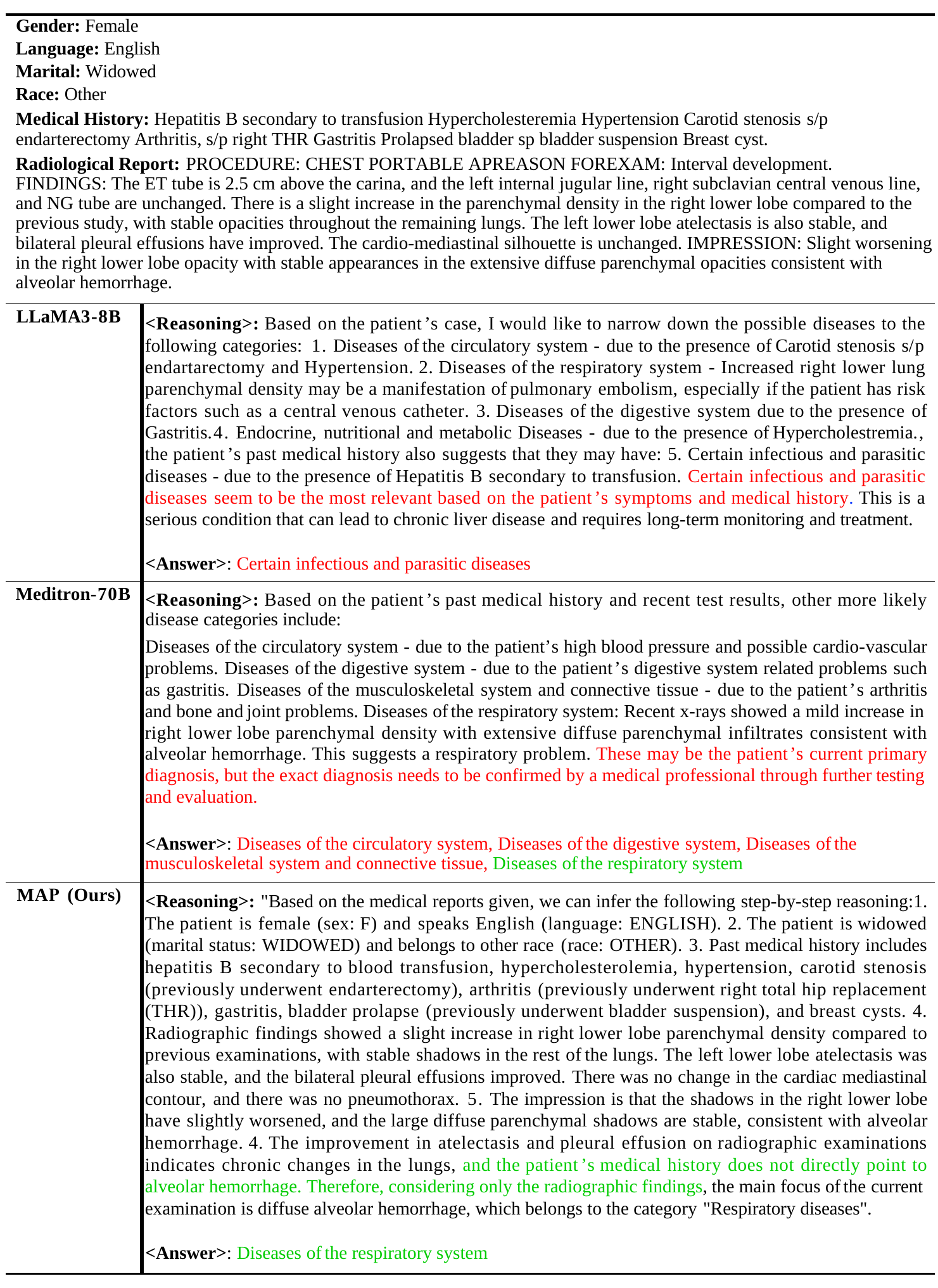}
\caption{The comparison shows the diagnosis and its reasoning of LLaMA3-8B, Meditron-70B, and our MAP. For better visualization, we highlight the incorrect parts in \textcolor{red}{red} and the correct parts in \textcolor{green}{green}.}
\label{fig: example}
\end{figure*}

\begin{figure*}[h]
\centering
   \includegraphics [width=0.88\textwidth]{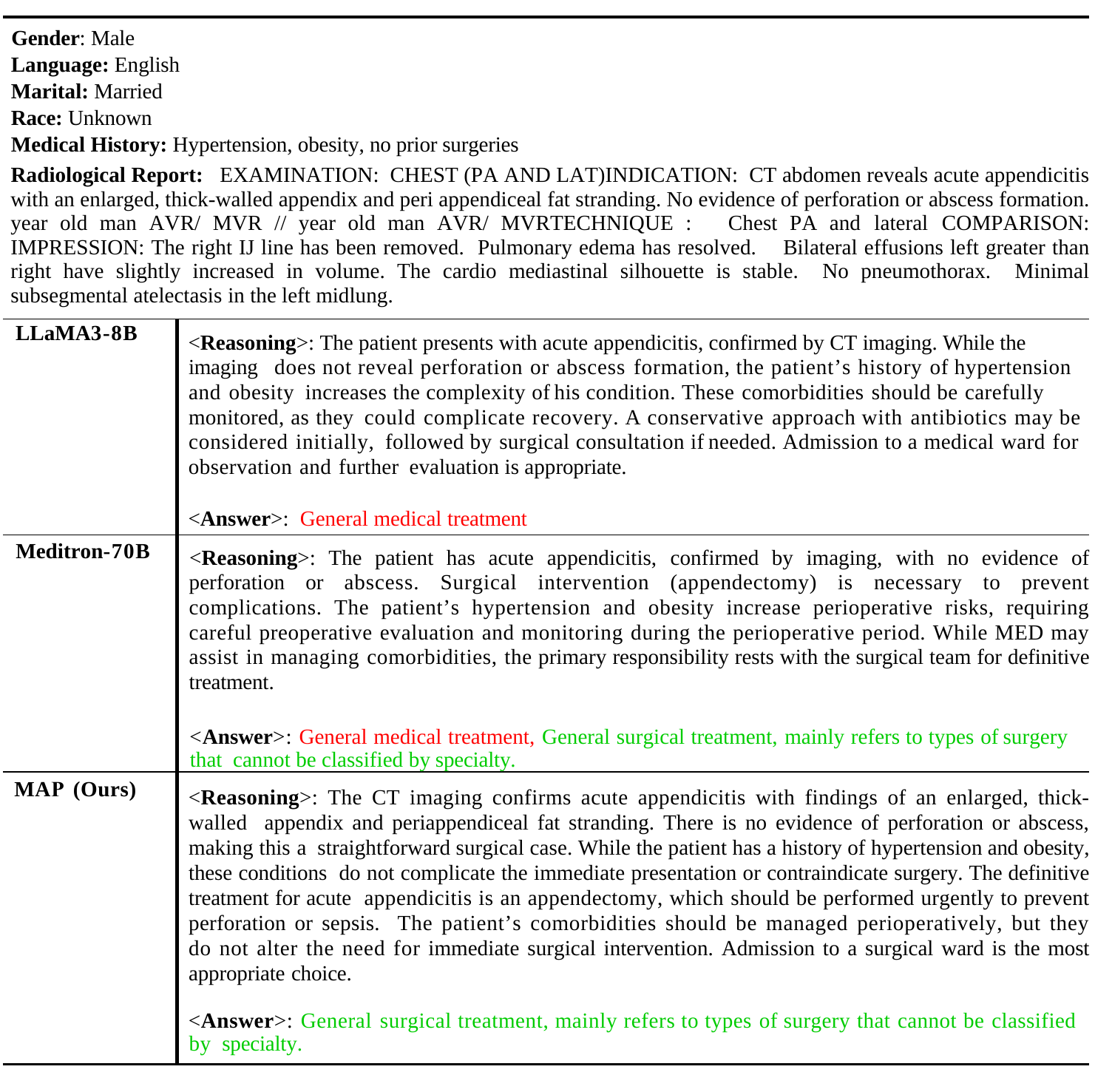}
\caption{The comparison shows the treatment and its reasoning of LLaMA3-8B, Meditron-70B, and our MAP. For better visualization, we highlight the incorrect parts in \textcolor{red}{red} and the correct parts in \textcolor{green}{green}.}
\label{fig: example3}
\end{figure*}

\noindent Comprehensive ablation studies in Figure \ref{fig: as2} were conducted to evaluate the module effectiveness across inpatient pathway tasks. We compared the performance in eight evaluation metrics and confirmed that the proposed MAP framework reveals consistent improvements over ablative baselines in most scenarios, as shown in Figure \ref{fig: example2} - \ref{fig: example3}. For instance, in Figure \ref{fig: example}, the baseline LLaMA-3-8B model tended to overemphasize medical history (e.g., infectious diseases) and overlooked critical radiographic findings (e.g., stable shadows in the lungs). In contrast, our MAP framework explicitly reasoned that the medical history has no direct correlation with current alveolar hemorrhage, demonstrating a structured and professional diagnostic approach. This structured reasoning process enabled the model to achieve more accurate diagnoses, particularly for respiratory diseases. The results highlight its ability to align reasoning patterns with professional clinical practice, effectively integrating diverse information sources into a coherent diagnostic framework. These systematic comparisons identified key imaging abnormalities and appropriately weighted different sources of information. 

\section{Data availability}
The dataset is available to all researchers who create an account on https://physionet.org/ and follow the steps to gain access to the MIMIC-IV database (https://physionet.org/content/mimiciv/3.0/). Access is given after being a credentialed user and completing the `CITI Data or Specimens Only Research' training course. The data use agreement of PhysioNet for the project must also be signed. The generated dataset can also be directly downloaded from PhysioNet. The corresponding data source is given in Table \ref{tab: data_source}.

\section{Code availability}
The evaluation framework used for this study can be found at GitHub (\href{https://github.com/franciszchen/MAP}{https://github.com/franciszchen/MAP}). The analysis framework to evaluate all results, generate all plots, and perform all statistical analysis can be found at Google Drive (\href{https://t.ly/N0JVT}{https://t.ly/N0JVT}). All code uses Python v3.10, PyTorch v2.1.2, Transformers v4.42.4, Spacy v3.7.5, LangChain v0.0.339, Llama-index-core v0.10.15, vllm v0.4.0+cu118, NLTK v3.8.1 and OpenAI v1.6.1. The code to create the dataset uses Python v3.10 and Pandas v2.1.3.

\section{Funding} 
This work was supported by the Hong Kong Research Grants Council (RGC) General Research Fund under Grant 14220622.

\section{Competing interests} 
The authors declare no competing interests.

\end{document}